\documentclass[preprint,journal]{vgtc}       % preprint (journal style)

\ifpdf%                                % if we use pdflatex
  \pdfoutput=1\relax                   % create PDFs from pdfLaTeX
  \usepackage{graphicx}                % allow us to embed graphics files
  \DeclareGraphicsExtensions{.pdf,.png,.jpg,.jpeg} % for pdflatex we expect .pdf, .png, or .jpg files
\else%                                 % else we use pure latex
  \usepackage{graphicx}                % allow us to embed graphics files
  \DeclareGraphicsExtensions{.eps}     % for pure latex we expect eps files
\fi%

\graphicspath{{figures/}{pictures/}{images/}{./}} % where to search for the images

\usepackage{microtype}                 % use micro-typography (slightly more compact, better to read)
\PassOptionsToPackage{warn}{textcomp}  % to address font issues with \textrightarrow
\usepackage{textcomp}                  % use better special symbols
\usepackage{mathptmx}                  % use matching math font
\usepackage{times}                     % we use Times as the main font
\usepackage{cite}                      % needed to automatically sort the references
\usepackage{tabu}                      % only used for the table example
\usepackage{booktabs}                  % only used for the table example
\usepackage{color}
\definecolor{red}{rgb}{0.8,0,0}
\definecolor{green}{rgb}{0.0,0.5,0}
\definecolor{blue}{rgb}{0.00,0.00,0.75}
\definecolor{gray}{rgb}{0.75,0.75,0.75}
\definecolor{orange}{rgb}{0.72,0.22,0.06}
\definecolor{purple}{rgb}{0.6,0.0,0.6}
\definecolor{darkred}{rgb}{0.6,0,0}
\definecolor{pink}{rgb}{0.58,0.12,0.3}

\newcommand{\equ}{equator}

\newcommand{\update}[1]{{#1}}

\preprinttext{}

\onlineid{1121}

\vgtccategory{Research}
\vgtcpapertype{theory/model}

\title{How do people explore virtual environments?}

\author{Vincent Sitzmann$^{*}$, Ana Serrano$^{*}$, Amy Pavel, \\Maneesh Agrawala, Diego Gutierrez, Belen Masia, Gordon Wetzstein}

\authorfooter{
\item
$^*$These authors contributed equally.
\item 
Correspondence to: sitzmann@cs.stanford.edu.
}

\abstract{Understanding how people explore immersive virtual environments is crucial for many applications, such as designing virtual reality (VR) content, developing new compression algorithms, or learning computational models of saliency or visual attention. Whereas a body of recent work has focused on modeling saliency in desktop viewing conditions, VR is very different from these conditions in that viewing behavior is governed by stereoscopic vision and by the complex interaction of head orientation, gaze, and other kinematic constraints. To further our understanding of viewing behavior and saliency in VR, we capture and analyze gaze and head orientation data of 169 users exploring stereoscopic, static omni-directional panoramas, for a  total of 1980 head and gaze trajectories for three different viewing conditions. We provide a thorough analysis of our data, which leads to several important insights, such as the existence of a particular fixation bias, which we then use to adapt existing saliency predictors to immersive VR conditions. In addition, we explore other applications of our data and analysis, including automatic alignment of VR video cuts, panorama thumbnails, panorama video synopsis, and saliency-based compression.
} % end of abstract


\CCScatlist{ % not used in journal version
 \CCScat{K.6.1}{Management of Computing and Information Systems}%
{Project and People Management}{Life Cycle};
 \CCScat{K.7.m}{The Computing Profession}{Miscellaneous}{Ethics}
}

\teaser{
  \centering
  \includegraphics[width=\textwidth]{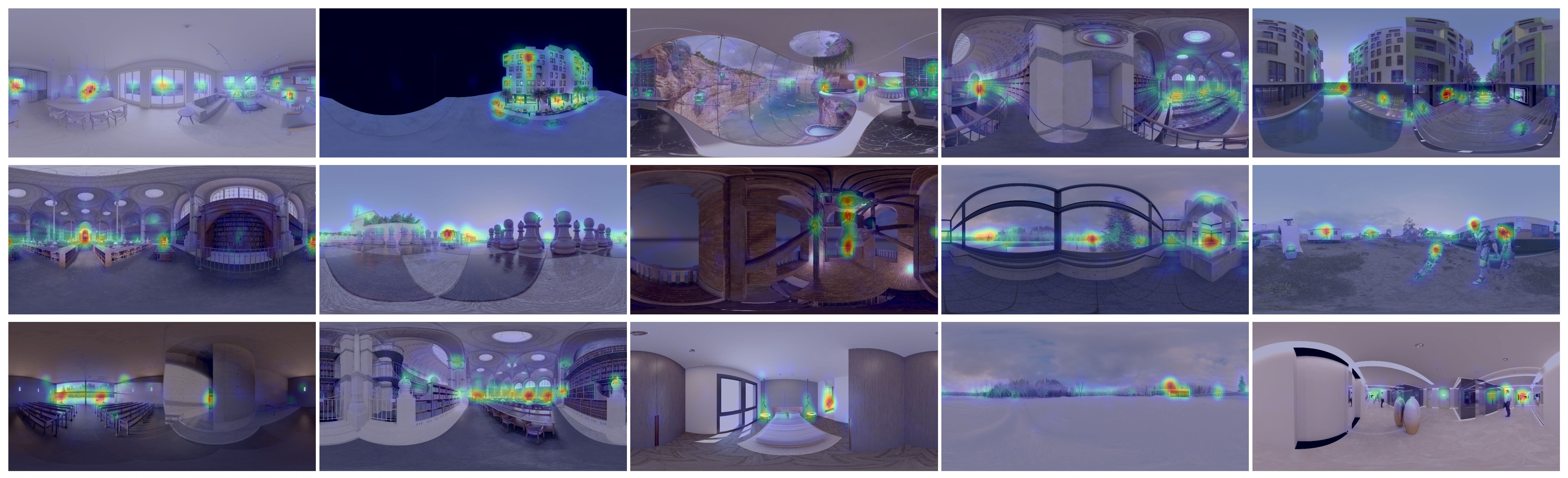}
  \caption{A representative subset of the 22 panoramas used to analyze how people explore virtual environments from a fixed viewpoint. We recorded almost two thousand scanpaths of users exploring these scenes in different immersive and non-immersive viewing conditions. We then analyzed this data, and provide meaningful insights about viewers' behavior. We apply these insights to VR applications, such as saliency prediction (shown in the image as overlaid heatmaps), VR movie editing, panorama thumbnail generation, panorama video synopsis, and saliency-aware compression of VR content.}
  \label{fig:mosaic}
	
}

\vgtcinsertpkg

\begin{document}

%% The ``\maketitle'' command must be the first command after the
%% ``\begin{document}'' command. It prepares and prints the title block.

%% the only exception to this rule is the \firstsection command
%\firstsection{Introduction}

\maketitle

\section{Introduction}
\label{Intro}

Virtual reality (VR) systems provide a new medium that has the potential to have a significant impact on our society. The experiences offered by these emerging systems are inherently different from radio, television, or theater, opening new directions in research areas such as cinematic VR capture~\cite{Anderson2016}, interaction~\cite{Sun2016}, or content generation and editing~\cite{Serrano2017,Nguyen:2017}. However, the behavior of users who visually explore immersive VR environments is not well understood, nor do statistical models exist to predict this behavior. Yet, with unprecedented capabilities for creating synthetic immersive environments, many important questions arise. How do we design 3D scenes or place cuts in VR videos? How do we drive user attention in virtual environments? Can we predict visual exploration patterns? How can we efficiently compress cinematic VR content? 

To address these and other questions from first principles, it is crucial to understand how users explore virtual environments. In this work, we take steps towards this goal. In particular, we are interested in quantifying aspects of user behavior that may be helpful in predicting exploratory user behavior \update{in static and dynamic virtual environments observed from a fixed viewpoint}. A detailed understanding of visual attention in VR would not only help answer the above questions, but also inform future designs of user interfaces, eye tracking technology, and other key aspects of VR systems.

%What makes VR different from conventional viewing conditions on desktop monitors is that viewing behavior in virtual environments is governed by the complex interaction of head orientation, gaze, and other kinematic constraints. Thus, 
A crucial requirement for developing an understanding for viewing behavior in VR is access to behavioral data. To this end, we have performed an extensive study, recording 1980 head and gaze trajectories from 169 people in 22 static virtual environments, which are represented as stereoscopic omni-directional panoramas. Data \update{is} recorded using a head-mounted display (HMD) in both standing and seated conditions (\emph{VR} condition and \emph{VR seated} condition), as well as for users observing the same scenes in mono on a desktop monitor for comparison (\emph{desktop condition}). 

\update{We analyze the data and discuss important insights (see Sec.~\ref{sec:Analysis} for more details).} We then leverage this to evaluate \update{\emph{existing}} saliency predictors, designed for narrow field of view video, in the context of immersive VR, and show how these can be adapted to VR applications. Saliency prediction is a well-explored topic and many existing models are evaluated by the MIT Saliency Benchmark~\cite{mitsaliencybenchmark}. 
%Given an image, these models predict a saliency map in the form of a 2D probability distribution which characterizes the probability that a location is fixated within a certain timeframe. 
However, these models assume that users sit in front of a screen while observing the images -- ground truth data is collected by eye trackers recording precisely this behavior. VR is different from traditional 2D viewing in that users naturally use both significant head movement and gaze to visually explore scenes; \update{we show that this leads} to a fixation bias that is not observed in conventional viewing conditions. Figure~\ref{fig:mosaic} shows panoramic views of some of our 22~scenes with superimposed saliency computed from the recorded scan paths in the VR condition. 
%Salient image regions in these stereoscopic 360$^\circ$ views match characteristics predicted by previous studies on saliency with conventional displays (e.g.,~\cite{Borji:2013,Bylinskii:2016,feng2016fixation}): people fixate consistently on faces, text, and vanishing points. However, view behavior in virtual environments is governed by a strong equator bias that is not observed in conventional viewing conditions. This bias is likely caused by the interaction of head orientation, gaze, and other kinematic constraints. Using the recorded data, we quantify the equator bias.
%
%
%\new{We provide a thorough statistical analysis of this data, including \diego{complete}. This analysis allows us to draw several conclusions. For instance, we have found a strong correlation with saliency predictions on conventional displays, and demonstrate how existing saliency models can be adapted to predict saliency for VR. However, view behavior in virtual environments is governed by the interaction of head orientation, gaze, and other kinematic constraints, which make these immersive scenarios more complex than observing a conventional display. \diego{complete}. These findings allows us to propose a series of example applications such as \diego{complete}
%}
%
Apart from saliency, we offer several other example applications that are directly derived from our findings. Specifically, our contributions are:
\begin{itemize}
	\item We record and provide an extensive dataset of visual exploration behavior in \update{stereoscopic, static omni-directional (ODS) panoramas}. The dataset contains head orientation and gaze direction, and it captures several different viewing conditions. All scenes, data, and code for analysis will be made public (Sec.~\ref{System}) 
	\item We provide low-level and high-level analysis of the recorded dataset. We derive relevant insights that can be crucial for predicting saliency in VR and other VR applications (Sec.~\ref{sec:Analysis})
	%One important insight provided by this analysis is a bias that quantifies users' preference to explore objects along the equator.
	\item We evaluate existing saliency predictors with respect to their performance in VR applications. We show how to tailor these predictors to \update{ODS panoramas} and we explore how useful saliency prediction from head movement alone is (Sec.~\ref{sec:prediction}) 
	%using the proposed equator bias.
	\item We demonstrate several applications of this saliency prediction, including automatic panorama thumbnails, VR video synopsis, compression, and VR video cuts (Sec.~\ref{sec:applications})
\end{itemize}

\section{Related Work}
\label{sec:related}

Modeling human gaze behavior and predicting visual attention has been an active area of vision research. In their seminal work, Koch and Ullman~\cite{Koch1987} introduced a model for predicting salient regions from a set of image features. Motivated by this work, many models of visual attention have been proposed throughout the last three decades. Most of these models are based on bottom-up, top-down, or hybrid approaches. Bottom-up approaches build on a combination of low-level image features, including color, contrast, or orientation ~\cite{Itti1998, Liu2011, Cheng2015, Kienzle2006} (see Zhao and Koch~\cite{Zhao2013} for a review). Top-down models take higher-level knowledge of the scene into account such as context or specific tasks~\cite{Jia2013,Judd2009,Goferman2012,Liu2014,Torralba2006}. Recently, advances in machine learning and particularly convolutional neuronal networks (CNNs) have fostered the convergence of top-down and bottom-up features for saliency prediction, producing more accurate models~\cite{Zhao2015,Wang2015,Pan2016,Li2015,huang2015salicon}. \update{Jiang et al.~\cite{jiang2015salicon} proposed a new methodology to collect attentional data on scales sufficient for these deep learning methods. Volotikin et al.~\cite{volokitin2016predicting} used features learned by CNNs to predict when saliency maps predicted by a model will be accurate and when fixations will be consistent among human observers.}
\update{Significant prior work explored rigorous benchmarking of saliency models, the impact of the metric on the evaluation result, and shortcomings of state-of-the-art models at the time~\cite{Bylinskii2016,Borji:2013,riche2013saliency}.}
%Work by Bylinskii et al.~\cite{Bylinskii2016} explored the shortcomings of state-of-the-art saliency models and provided a rigorous base for saliency map benchmarking. 
Recent work also attempts to extend CNN approaches beyond classical 2D images by computing saliency in more complex scenarios such as stereo images~\cite{Guo2014,Cong2016} or video~\cite{Chaabouni2016,Leifman2016}. A related line of research is devoted to modeling the gaze scanpath followed by subjects, i.e., the temporal evolution of the viewer's \update{gaze~\cite{LeMeur2015, jiang2016learning}}. \update{Marmitt et al.~\cite{marmitt2002modeling} developed a metric to evaluate predicted scanpaths in VR and showed that predictors built for classic viewing conditions perform significantly worse in VR.} 
Building on the rich literature in this area, we explore user behavior and visual attention in immersive virtual environments, which can help build similar models for VR.

%A related line of research is devoted to modeling not only the stationary, final saliency map, but the gaze scanpath followed by subjects, i.e., the temporal evolution of the viewer's gaze~\cite{LeMeur2015}. These works typically provide statistical models based on prior knowledge of visual behavior; our gathered data and analysis can help in building similar models for VR 360$^\circ$ content.

What makes VR different from desktop viewing conditions is the fact that head orientation is used as a natural interface to control perspective \update{(and in some cases navigation as well~\cite{Tregillus2017})}. The interactions of head and eye movements are complex and neurally coupled, for example via the vestibulo-ocular reflex~\cite{Laurutis1986}. \update{Koehler et al.~\cite{koehler2014saliency} showed that saliency maps can differ depending on the instructions given to the viewer.} For more information on user behavior in VR, we refer to Ruhland et al.~\cite{Ruhland2015}, who provide a review of eye gaze behavior, and Freedman~\cite{Freedman2008}, who discusses the mechanisms that characterize the coordination between eyes and head during visual orienting movements. With the data recorded in this project, we observe the vestibulo-ocular reflex and other interesting effects. In the paper and supplemental material, we provide an extensive analysis of the user data and derive statistics modeling many low-level aspects of viewing behavior. We hope that this analysis will be useful for basic vision research.

%More recently, Serrano et al.~\shortcite{Serrano2017} conducted a series of experiments to analyze continuity editing in VR video; to do so, they proposed a number of metrics to assess viewers' attentional behavior in VR environments, some of which we use in this work. 

Recent work of Nakashima et al.~\cite{Nakashima2015} is closely related to some aspects of our work. They propose a head direction prior to improve accuracy in saliency-based gaze prediction through simple multiplication of the gaze saliency map by a Gaussian head direction bias. The data collected in this paper and in-depth analyses augment prior work in this field, and may allow for future data-driven models for visual behavior to be learned. 

%Also, Kollenberg et al.~\shortcite{Kollenberg2010} propose a basic analysis of head-gaze interaction in VR. 

Finally, gaze tracking has found many applications in VR user interfaces~\cite{Tanriverdi2000} and gaze-contingent displays~\cite{Duchowski:2004,stengel2016compdisp,Padmanaban:2017:GazeContingentFocus}. The ability to predict viewing behavior would be helpful for all of these applications. For example, gaze-contingent techniques may become possible without dedicated gaze trackers, which are currently expensive and not widely available. Moreover, techniques for editing VR content are starting to emerge~\cite{Nguyen:2017,Serrano2017}. The understanding of user behavior we aim to develop in this paper could also influence these and other tools for content creation. 

A preliminary version of this manuscript was published on arXiv \cite{SaliencyVR:2016}.

\section{Recording Head Orientation and Gaze }
%\section{Dataset for the study of human behavior in VR}
\label{System}

In this section, we summarize our efforts towards recording a dataset that contains head orientation and gaze direction for users watching stereoscopic VR panoramas in several different viewing conditions; we provide additional details in the supplemental material. These data form the basis of a statistical analysis of viewing behavior (Sec.~\ref{sec:Analysis}), as ground truth for saliency prediction (Sec.~\ref{sec:prediction}), and also as reference saliency for several higher-level applications (Sec.~\ref{sec:applications}). 
%All data described in this section will be made publicly available along with code to process it.

\subsection{Data capture}

\paragraph{Stimuli} For the experiments reported in this paper, we used 22 high-resolution omni-directional stereo panoramas (see Figure~\ref{fig:mosaic} and supplemental material). \update{We opt for a fixed viewpoint because for the subsequent analyses it is crucial that subjects see the exact same content; further, in a 3D scenario the variability is likely to be much higher, requiring extremely large numbers of subjects to draw significant conclusions.} The scenes include \update{(14)} indoor and \update{(8)} outdoor scenarios and do not contain landmarks that may be recognized by the users. \update{For each scene we explore different conditions, which limits the number of scenes we can have with the experiment size remaining tractable; with the current stimuli and conditions we have collected nearly 2,000 trajectories from 169 viewers.} All scenes are computer generated by artists; we received permission to use them for this study.
%To choose a set of stimuli, we take into account several criteria: (i) \textit{familiarity:} the scenes depicted need to be familiar to the viewers, in order to avoid confusion which could translate into erratic gaze behavior; (ii) \textit{recognition:} participants should not recognize the scenes (for instance, a famous landmark), to avoid knowledge from previous experience interfering with such gaze behavior; (iii) \textit{artifact-free:} they should be free of obvious artifacts that may draw unwanted attention and thus affect gaze; and (iv) \textit{coverage:} the stimuli need to cover a wide enough set of scenarios, both indoor and outdoor. Based on these criteria, we selected a final set of 22 omni-directional stereo panoramas (shown in Fig.~\ref{fig:mosaic} and in the supplemental material). All scenes are computer-generated by artists; we received permission to use them for this study. 

\paragraph{Conditions}
We recorded users observing the 22 panoramas under three different conditions: 
%\update{that differ on how the user can orient the viewport to explore the scene}: 
in a standing position using a head-mounted display (i.e., the \emph{VR condition}), seated in a non-swivel chair using a head-mounted display (i.e., the \emph{VR seated condition}\update{, making it more difficult to turn around}), and seated in front of a desktop monitor (i.e., the \emph{desktop condition}). In the desktop condition, the scenes are monoscopic, and users navigate with a mouse. For each scene, we tested four different starting points, spaced at 90$^\circ$ longitude, which results in a total of 264 conditions. These starting points were chosen to cover the entire longitudinal range, while keeping the number of different conditions tractable. \update{We chose not to randomize the starting point over the whole latitude (and rather select randomly from four fixed ones) to limit the number of conditions while being able to analyze the influence of the starting point (Sec.~\ref{subsec:viewport} and supplement); complete randomization over the starting point could be of interest for future studies.}
%
%\new{ADD VR seated condition}

\paragraph{Participants}
For the \emph{VR} condition, we recorded 122 users (92 male, 30 female, age 17-59). The experiments with the \emph{VR seated} condition were performed by 47 users (38 male, 9 female, age 17-39). Users were asked to first perform a stereo vision (Randot) test to quantify their stereo acuity. For \emph{desktop} experiments, we recruited 44 additional participants (27 male, 17 female, age 18-33). All participants reported normal or corrected-to-normal vision. 

\paragraph{Procedure}
All VR scenes were displayed using an Oculus DK2 head-mounted display, equipped with a pupil-labs\footnote{\url{https://pupil-labs.com}} stereoscopic eye tracker recording at 120~Hz. The DK2 offers a field of view of $95 \times 106^\circ$. 
%Prior to displaying test scenes, the eye tracker was calibrated using a per-user, per-use procedure. 
The Unity game engine was used to display all scenes and record head orientation while the eye tracker collected gaze data on a separate computer. Users were instructed to freely explore the scene and were provided with a pair of earmuffs to avoid auditory interference. 
Scenes and starting points were randomized, while ensuring that each user would only see the same scene once from a single random starting point. Each user was shown 8 scenes;  \update{each scene in a certain condition was shown to the user during 30 seconds, while the total time per user that the experiment took,} including calibration \update{and explanation}, was approximately 10 minutes.

\update{We modeled the \emph{desktop} condition after typical, mouse-controlled desktop panorama viewers on the web (i.e., YouTubeVR or Facebook360).} Users sat 0.45 meters away from a 17.3'' monitor with a resolution of $1920 \times 1080$~px, covering a field of view of $23 \times 13^\circ$. We used a Tobii EyeX eye tracker with an accuracy of 0.6$^\circ$ at a sampling frequency of 55~Hz~\cite{gibaldi2016evaluation}. The image viewer displayed a rectilinear projection of a $97 \times 65^\circ$ viewport of the panorama. \update{To keep the field of view consistent, no zooming was possible.} We instructed the users on how to use the image viewer, before showing the 22 scenes for 30 seconds each. In this condition, we only collected gaze data since users rarely re-orient their head. Instead, we recorded where the users interactively place the virtual camera in the panorama as a proxy for head orientation.

\subsection{Data processing}

To identify fixations, we transformed the normalized gaze tracker coordinates to latitude and longitude in the 360$^\circ$ panorama. This is necessary to detect users fixating on panorama features while turning their head. We used thresholding based on dispersion and duration of the fixations~\cite{Salvucci2000}. For the VR experiments, we set the minimum duration to 150~ms~\cite{Salvucci2000} and the maximum dispersion to 1$^\circ$~\cite{Blignaut2009}. For the desktop condition, we first smoothed this data with a running average of 2 samples, and detected fixations with a dispersion of 2$^\circ$. We counted the number of fixations at each pixel location in the panorama. Similar to Judd et al.~\cite{Judd2009},  we only consider measurements from the moment where user's gaze left the initial starting point to avoid adding trivial information. We convolved these fixation maps with a Gaussian with a standard deviation of 1$^\circ$ of visual angle to yield continuous saliency maps~\cite{LeMeur2013}.
 
\section{Understanding viewing behavior in VR}
%\section{Analysis}
\label{sec:Analysis}

With \update{the} recorded data, we can gather insights and investigate a number of questions about the behavior of users exploring virtual environments. In the following, we analyze both low-level characteristics, such as duration of the fixations and speed of gaze, and higher-level characteristics, such as the influence of the content or characteristics of the scene. 
%The primary goal of this analysis is to understand and quantify user behavior in VR.

%This analysis serves a two purposes: first, it can help in the creation of gaze scanpath predictors or saliency predictors for VR. Second, it provides guidelines for content creators on how to indirectly control user attention. We structure this section according to the different questions that our data can help answer. %In this section, we first analyze look at the low-level data (Section~\ref{subsec:low}), and then move on to the higher-level analysis (Section~\ref{subsec:high}). }

\subsection{Is viewing behavior similar between users?}
\label{subsec:agreement}
\begin{figure}[t]
\includegraphics[width=\columnwidth]{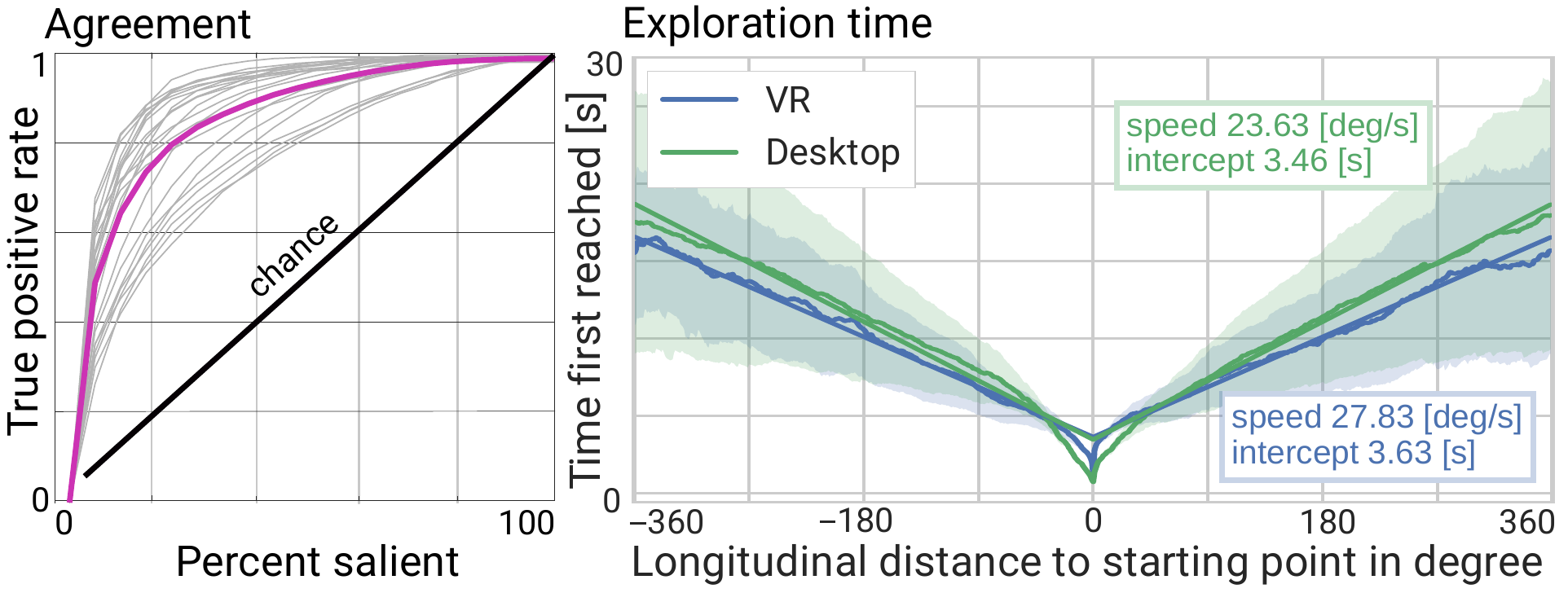} 
\caption{\label{fig:ExpTime_Agreement} \emph{Left:} ROC curve of human performance averaged across users (magenta) and individual ROCs for each scene (light gray). The fast convergence to the maximum is indicative of a strong agreement between users. \emph{Right:} Exploration time computed as the average time until a specific longitudinal offset from the starting point is reached.}
\end{figure}

We first want to assess whether viewing behavior between users is similar; this is also indicative of how robust our data is, and thus how much we can rely on it to draw conclusions. To answer this, we analyze the agreement between users. Specifically, we compute the \emph{inter-observer congruency} metric by means of a \emph{receiver operating characteristic} curve (ROC)~\cite{Torralba2006,LeMeur2013}. This metric calculates the ability of the $i^{th}$ user to predict a \emph{ground truth saliency map}, which is computed from the fixations of all the other users averaged. A single point in the ROC curve is computed by finding the top $n$\% most salient regions of the ground truth saliency map (leaving out the $i^{th}$ user), and then calculating the percentage of fixations of the $i^{th}$ user that fall into these regions. We  show the average ROC for all the 22 scenes in Figure~\ref{fig:ExpTime_Agreement} (left), compared with chance (the individual ROCs for each scene are depicted in light gray). The fast convergence of these curves to the maximum rate of 1 indicates a strong agreement, and thus similar behavior, between users for each of the scenes tested. 70\% of all fixations fall within the 20\% most salient regions. These values are surprisingly good, since they are comparable to previous studies viewing regular images on a display~\cite{LeMeur2013}. 
%Thus, viewing behavior is indeed similar between users.

%

\subsection{How different is viewing behavior for the 3~conditions?} 
\label{subsec:compare}
An important question to ask is whether viewing behavior changes when exploring a scene under different conditions. Visual inspection of our three conditions (\emph{VR}, \emph{VR seated}, and \emph{desktop}) shows a high similarity between the saliency maps (see supplement). For a quantitative evaluation of the similarity of saliency maps (here, and in the rest of the paper), we use the Pearson correlation (CC) score, which is a widely used metric in saliency map prediction~\cite{Bylinskii2016}. The high similarity is confirmed by a median CC score of $0.80$ when comparing the \emph{VR} and the \emph{VR seated} conditions, and $0.76$ when comparing the \emph{VR} and the \emph{desktop} conditions. The latter is a significant insight: since desktop experiments are much easier to control, it may be possible to use these for collecting adequate training sets for data-driven saliency prediction in future VR systems. Given this similarity, we report only the results of the \emph{VR} (standing) condition throughout the remainder of the paper, unless a significant difference is found, and refer the reader to the supplemental for the \emph{VR seated} and \emph{desktop} conditions. 
%An important question to ask is whether viewing behavior changes when exploring a scene with a HMD, or on a desktop with a mouse \diego{these conditions must have been clearly described in advance}. 
%%If the behavior is similar, this would mean that a simple desktop environment can be used to collect saliency maps that are a very close to the VR condition saliency maps. 
%Our data shows in fact a very high similarity between the two viewing conditions. This is encouraging: since desktop experiments are much easier to control, this insight makes it more feasible to collect training sets for data-driven saliency prediction in future VR systems. Given this finding, throughout the paper we report only the results of the VR viewing condition, unless a significant difference is found, and refer the reader to the supplemental for the desktop condition. 
%\diego{place elsewhere? We now talk about saliency maps and IOVCs, too early....}
%%
%We compare saliency maps obtained from the VR and desktop experiments. Visual inspection shows a high similarity between the saliency maps (see supplement). This is confirmed by a high median CC score of $0.79$, which is computed by averaging over all per-scene pairwise CC scores between saliency maps of the two conditions. 

%\subsection{Quantifying the equator bias for fixations in VR}
\subsection{Is there a fixation bias in VR?}
\label{subsec:bias}
\begin{figure}[t]
\includegraphics[width=\columnwidth]{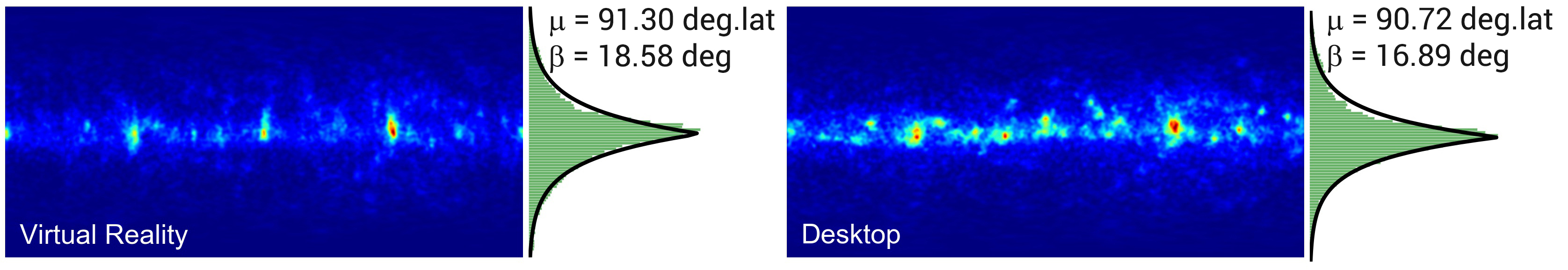} 
\caption{\label{fig:eqbias} Average saliency maps computed with all the scenes for both the \emph{VR} (left) and the \emph{desktop} (right) conditions. These average maps demonstrate an ``\equ~bias'' that is well-described by a Laplacian fit modeling the probability of a user fixating on an object at a specific latitude.}
\end{figure}

Several researchers have reported a strong bias for human fixations to be near the center, when viewing regular images~\cite{Judd2009,Nuthmann2010}. A natural question to ask is whether a similar bias exists in VR. Similar to Judd et al.~\cite{Judd2009}, we calculate the average of all 22 saliency maps, and filter out fixations within the close vicinity (20$^\circ$ longitude) of the starting point. The resulting data indicates that users tend to fixate around the \equ~of the panoramas, with very few fixations in latitudes far from it. To quantify this \emph{\equ~bias}, we marginalize out the longitudinal component of the saliency map, and fit a Laplace distribution---with location parameter $\mu$ and diversity $\beta$---to the latitudinal component (this particular distribution yielded the best match among several tested distributions). Figure~\ref{fig:eqbias} depicts the average saliency map, as well as our Laplacian fit to the latitudinal distribution and its parameters, for both the \emph{VR} and the \emph{desktop} conditions. While the mean is almost identical, the \equ~bias for the desktop condition has a lower diversity. As discussed in Section~\ref{sec:prediction}, this Laplacian equator bias is crucial for predicting saliency in VR. 

Note that most of the scenes in our study have a clear horizon line, which may have influenced the observed equator bias along with viewing preferences, kinematic constraints, \update{as well as the static nature of the scenes}. \update{However, most virtual environments share this type of scene layout, so we believe our findings generalize to a significant fraction of this type of content.}
\update{Further, even for scenes with content scattered along different latitudes (see, e.g., \emph{scene 16} in Fig. 12 of the supplement, displaying very few salient areas near the poles), we observed an equator bias. Nevertheless, different tasks or scenarios, such as gaming, may influence this bias.}

%

%In most of the regular images, the most important information is placed away from the poles, which may have an influence in the reported center bias. We note that most of our scenes have a clear horizon line, along which most relevant objects of the scenes are naturally located;  this may be a factor for the observed equator bias. Nevertheless, such horizon line is shared by most virtual environments, to act as a reference and mimic the real world. 

\subsection{Does scene content affect viewing behavior?}
\label{subsec:entropy}
A fundamental issue when analyzing viewing behavior is the potential influence of scene content. This is of particular relevance for content creators; since in a VR setup the viewer has control over the camera, this analysis can help address the key challenge of predicting user attention. 
%Can we guide her gaze scanpath when exploring the scene? Can we make her explore it faster or more carefully, fixating on details?

\begin{figure}[t]
\includegraphics[width=\columnwidth]{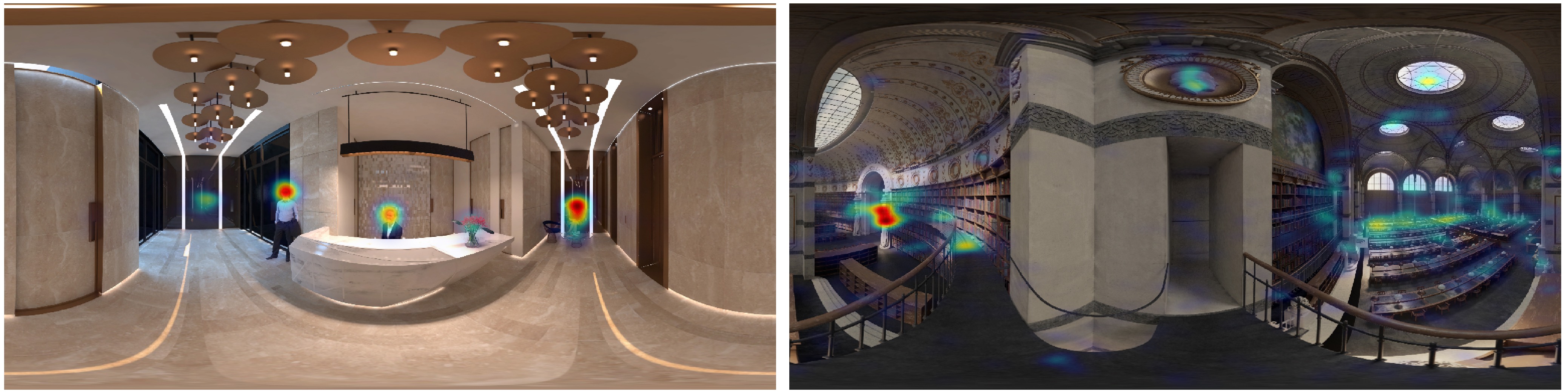} 
\caption{\label{fig:entropy}Saliency maps presenting the lowest (\emph{left}) and highest (\emph{right}) entropy in our dataset. Saliency maps with low entropy have very defined salient regions while in maps with high entropy fixations are scattered all over the scene.}
\end{figure}

To characterize scene content in a manner that enables \update{insightful analysis}, we rely on the distribution of salient regions in the scene, in particular on the \emph{entropy} of the saliency maps. A high entropy results from a large number of similarly salient objects distributed throughout the scene, causing users' fixations to be scattered all over the scene; a low entropy results from a few salient objects that capture all the viewer's attention. Figure~\ref{fig:entropy} shows the saliency maps of the scenes with lowest and highest entropy in our dataset.

Our entropy is computed as the Shannon entropy of the \emph{ground truth saliency map}, computed, per scene, from the average of all users~\cite{Judd2009}. The entropy is given by:  $- \sum_{i=1}^{N} s_i^2 log(s_i^2)$, with $s$ being the ground truth saliency, and $N$ the number of pixels. We consider two entropy levels, low and high, which we term $\{E_{0}, E_{1}\}$, respectively. Since a clear threshold for classifying each scene according to its entropy does not exist, we take a conservative approach and analyze only the four scenes with highest and the four with lowest entropy, for a total of eight scenes. 
%, and then normalized. 
% 

\subsubsection{Viewer's behavior metrics}
\label{subsubsec:metrics}
\begin{figure}[t]
\includegraphics[width=0.49\columnwidth]{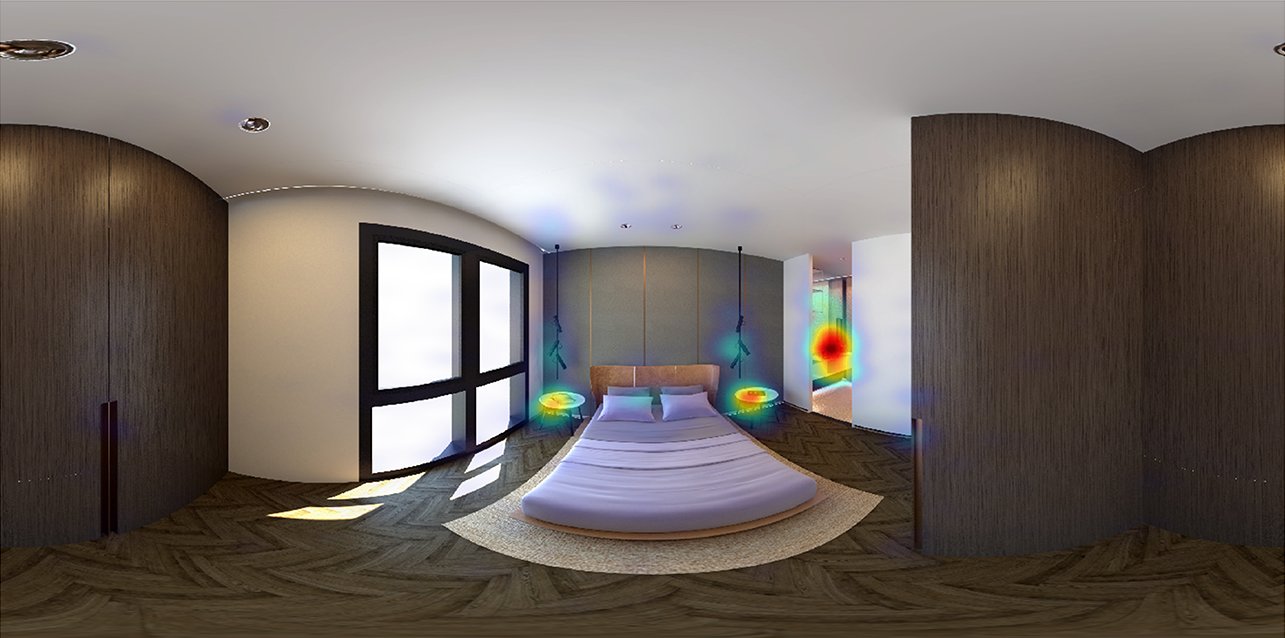} 
\includegraphics[width=0.49\columnwidth]{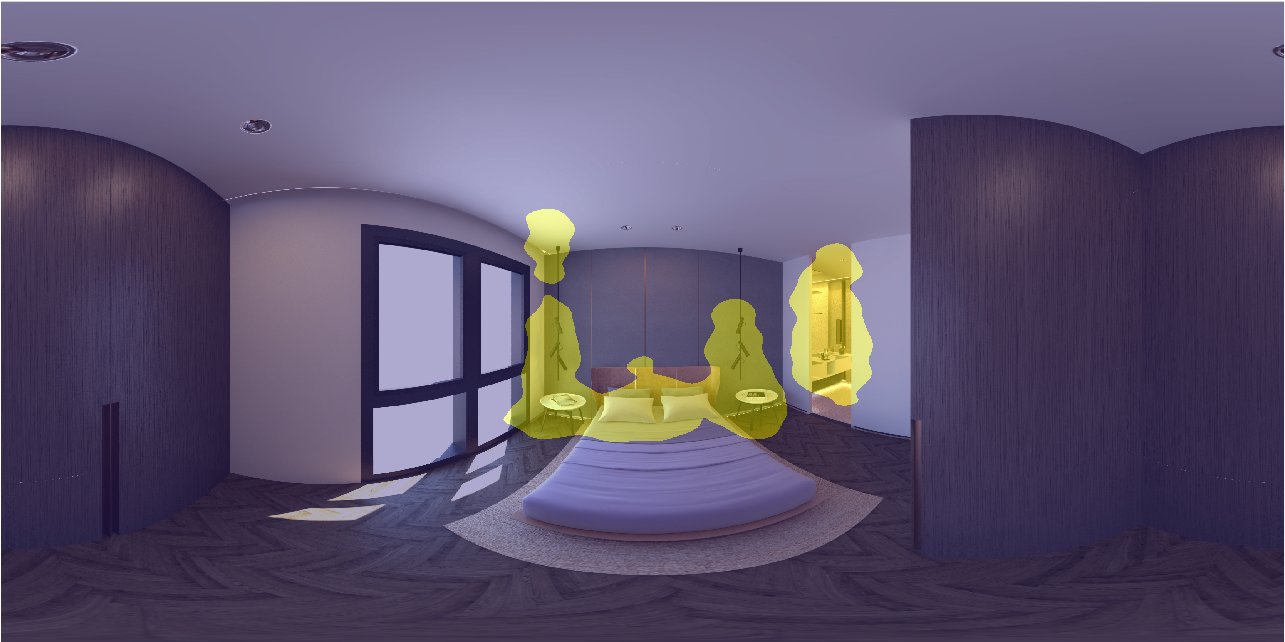} 
\caption{\label{fig:5percentROI} Salient region computation. \emph{Left:} Ground truth saliency map for a sample scene. \emph{Right:} Corresponding salient regions (yellow) computed by thresholding the 5\% most salient pixels of the scene.}
\end{figure}

Measuring viewer's behavior in an objective manner is not a simple task. First, we define \textit{salient regions} as the 5\% most salient pixels of a scene. Figure~\ref{fig:5percentROI} shows a saliency map and the resulting salient regions computed with this criterion. We then rely on three metrics recently proposed by Serrano et al.~\cite{Serrano2017} in the context of gaze analysis for VR movie editing (time to reach a salient region (\emph{\textmd{timeToSR}}), percentage of fixations inside the salient regions (\emph{\textmd{percFixInside}}), and number of fixations (\emph{\textmd{nFix}}), which are summarized in the supplemental material), and propose a fourth, novel one, tailored for static 360$^{\circ}$ panoramas:

%\paragraph{Time to reach a salient region (\emph{\textmd{timeToSR}})} It accounts for the time before the user fixates on a salient region. 
%%
%\paragraph{Percentage of fixations inside the salient regions (\emph{\textmd{percFixInside}})} It is computed after the user fixates on a salient region for the first time, in order to make it independent of \emph{timeToSR}. It is indicative of the interest of the user in the salient regions.
%%
%\paragraph{Number of fixations (\emph{\textmd{nFix}})} Computed as the ratio between the number of fixations and the number of gaze samples. 
%%
\paragraph{Convergence time (\emph{convergTime})} For every scene, we obtain the per-user saliency maps at different time steps, and compute the similarity (CC score) with the fully-converged saliency map. We plot the temporal evolution of this CC score, and compute the area under this curve. This metric represents the temporal convergence of saliency maps; it is inversely proportional to how long it takes for the fixation map during exploration to converge to the ground truth saliency map.

\subsubsection{Analysis}
\label{subsubsec:analysis}
We first test for independence of observations performing a Wald's test (please refer to the \update{supplement}). Based on its results, we employ ANOVA when analyzing \emph{percFixInside}, since the samples are considered to be independent, and report significance values obtained from multilevel modeling for the other three metrics. 

%The design of our experiment implies that the same user can see several (but not all) of the tested conditions, so we first need to test for independence of observations (Wald's test). For one of the metrics (\emph{percFixInside}) the effect of the user was found to be not significant ($p=0.071$), while for the other three (\emph{timeToSR} and \emph{nFix}, and \textit{convergTime}) it was significant ($p=0.023$, and $p<0.001$, and $p<0.001$, respectively). We therefore employ ANOVA when analyzing \emph{percFixInside}, since the samples are considered to be independent, and report significance values obtained from multilevel modeling~\cite{Browne2004,Raudensbush2002} (which is well-suited for related data) for the other three metrics. 
%We include here results of the analysis for the standing VR experiment, while the results for the desktop setup can be found in the supplemental. The results of the desktop analysis are nevertheless consistent with those reported for the VR condition.} \belen{Check. Also, say sth. about seating?

We find that the \emph{entropy} of the scene has a significant effect on \textit{nFix} ($p<0.001$), \textit{timeToSR} ($p<0.001$), \textit{percFixInside} ($p=0.022$), and \textit{convergTime} ($p<0.001$). Specifically, on scenes with low entropy ($E_{0}$), the time to reach a salient region (\emph{timeToSR}) is lower. This may be counter-intuitive, since high entropy scenes contain a larger number of salient regions and thus it would be easier to reach one; interestingly, though, our results indicate that the viewer explores the scene faster in cases of low entropy, quickly discarding non-salient regions, and that her attention gets directed towards the few salient regions faster. This hypothesis is further supported by the behavior of the \textit{convergTime} metric, which shows that scenes with low entropy do converge faster, and is consistent with the number of fixations, and fixations inside the salient region (\emph{nFix} and \emph{percFixInside}): both are higher for low entropy scenes, indicating that users pay more attention to salient regions when such regions are less, and more concentrated.

\subsection{Does the starting point affect viewing behavior?}
\label{subsec:viewport}
%Another important question, related to the previous one, is whether the particular point (viewport) at which the viewer starts exploring the scene has an influence when exploring the panorama. To make this analysis tractable, we have tested four equally-spaced \textit{viewports}\footnote{Viewports match the FOV of the Oculus,  97$^{\circ}$. Thus the four viewports define slightly overlapping areas.} per scene, covering the 360$^{\circ}$. 
%%However, each of the four viewports $\{1..4\}$ may have a different meaning, depending on the scene. We therefore need to recategorize our data, to be able to provide a meaningful analysis of this factor across scenes. 
%We define two possible values for each viewport, $\{V_{0}, V_{1}\}$, corresponding to whether or not they contain salient regions.
%%, and use the previously defined metrics (Section~\ref{subsubsec:metrics}) for our analysis.
%%
%We find that the \emph{viewport} condition has a significant influence on \emph{timeToSR}, which is not surprising ($p<0.001$): If it does not contain salient regions ($V_{0}$), viewers take longer to reach them elsewhere. There is also a significant influence on \textit{convergTime} ($p=0.004$); specifically, users converge faster towards the final saliency map when the initial viewport contains salient regions ($V_{1}$). 
%%\belen{We need to check if this is true even if considering only cases of high entropy, this is what would be interesting.}. 
%There is no influence of the viewport on the number of fixations, nor in the number of fixations inside salient regions. 
%
We also evaluate whether the starting viewport conditions the final saliency map for a given scene: For each scene, we compute the similarity between the final saliency map of the $i^{th}$ viewport and the other three, using again the CC score. We obtain a median CC score of $0.79$, which indicates that the final saliency maps after 30 seconds, starting from different viewports, converge and are very similar.  
%(see Section~\ref{subsec:agreement}).} 
%
Additional analysis on the influence of the viewport, including also a state sequence analysis~\cite{TraMineR,Serrano2017}, can be found in the \update{supplement}.
%We have performed an additional \emph{state sequences} analysis~\cite{TraMineR,Serrano2017}, which confirms these findings; please refer to the supplement for this information. 

\subsection{How are head and gaze statistics related?}
\label{subsec:lowlevel}

%Predicting gaze scanpaths of observers when freely exploring a VR panorama would be very interesting in many fields, including vision, cognition, and of course, any VR-related application. Since the seminal work of Koch and Ulman~\shortcite{Koch1987}, many researchers have proposed models of human gaze when viewing regular 2D images on conventional displays (e.g.,~\cite{LeMeur2015, Wang2011, Hacisalihzade1992, Boccignone2004}). An important element to derive such models is gaze statistics, and whether those found in our VR setup are comparable to the ones reported for traditional viewing conditions; this would inform to what extent we can use existing gaze predictors in VR applications, or be useful as priors in the development of new predictors. Moreover, we analyze potential interactions between head and gaze movement; this can be of particular interest to build gaze predictors using just head movement as input, since head position is much cheaper to obtain than actual gaze data. 
Many additional insights can be learned from our data, which may be useful for further vision and cognition research, or in applications that require predicting gaze or saliency in VR (see also Section~\ref{sec:prediction}).
First, we evaluate the speed with which users explore a given scene. Figure~\ref{fig:ExpTime_Agreement} (right) shows this \emph{exploration time}, which is the average time that users took to move their eyes to a certain longitude relative to their starting point. On average, users fully explored each scene after about 19 seconds.

In our experiments, the mean number of fixations across scenes is 50.35 $\pm$ 14.63, 
% and 40.04 $\pm$ 13.36 for the desktop condition. 
while the duration of these fixations is 257~ms $\pm$ 121.
%For the desktop condition, we measure 245~ms $\pm$ 114. 
This is in the range reported for traditional screen viewing conditions~\cite{Salvucci2000}. The mean gaze direction relative to the head orientation across scenes is 13.85$^\circ$ $\pm$ 11.73, which is consistent with the analysis performed by Kollenberg et al.~\cite{Kollenberg2010}.
%, who reported narrower eye-rotations for HMD than for traditional viewing conditions. 
%\diego{the decision was to remove the table; but I've kept this here since Belen left a note...}
% In table Table~\ref{tab:speeds}, we show the mean speed for gaze and head movements (longitudinal and latitudinal) for VR, as well as the mean gaze speed for the desktop condition. Interestingly, both head and gaze move much slower in the latitudinal direction. 
%\belen{We need to say something more about speed than: there you have the table - but let's wait for the final numbers.} 

\begin{figure}[t]
\includegraphics[width=\columnwidth]{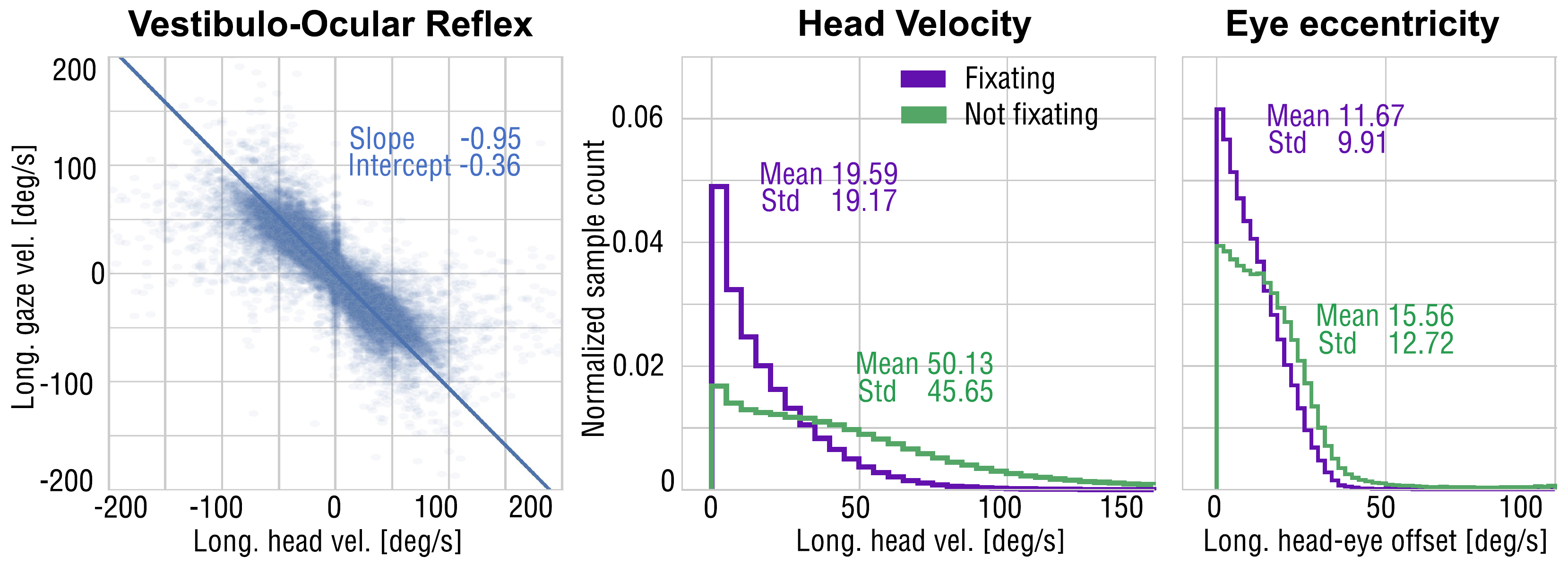} 
\caption{\label{fig:head_fixation_vor}\emph{Left:} the vestibulo-ocular reflex demonstrated by an inverse linear relationship of gaze and head velocities. \emph{Middle and right:} distributions of longitudinal head velocity and longitudinal eye eccentricity, respectively, while fixating and while not fixating.}
\end{figure}

We have also identified the \emph{vestibulo-ocular reflex}~\cite{Laurutis1986} in our data. This reflexive mechanism moves the eyes contrary to the head movement, in order to stabilize the line of sight and thus improve vision quality.  Figure~\ref{fig:head_fixation_vor} (left) shows the expected inverse linear relationship between head velocity and relative gaze velocity when fixating. 
Given this observation, we further analyze 
%interactions between eye and head rotation (latitude and longitude). In particular, to analyze 
the interaction between eye and head movements when shifting to a new target.
% we look for correlations in the acceleration of both gaze and head. To do this, 
 We offset in time head and gaze acceleration measurements relative to each other, and compute the cross-correlation for different temporal shifts. Our data reveals that head follows gaze with an average delay of 58~ms, where the largest cross-correlation is observed, consistent with previous works~\cite{Freedman2008, Doshi2012}.

%, reporting delays in head movement when shifting to a non-predictable target (i.e., not trained nor premeditated gaze shifts). 
%

%\subsection{Are there different viewing modes in VR?}
%\label{subsec:modes}
It is well-known that gaze velocities differ when users fixate and when they do not~\cite{Salvucci2000}. We look at whether this is also the case for head velocities, since they could then act as a rough proxy for fixation classification. 
%We also found that the statistics of head and eye movements differ when users fixate versus when they do not fixate. 
Figure~\ref{fig:head_fixation_vor} (middle) shows that users move their head at longitudinal velocities significantly below the average head speed when they are fixating, and above average when they are not. 
%\diego{remove?: }The same effect, but to a lesser degree, can be seen in the latitudinal head velocity (see supplement). 
Further, Figure~\ref{fig:head_fixation_vor} (right) shows that the longitudinal rotation angle of the eyes relative to the head orientation (eye eccentricity) is significantly smaller when users are fixating.  
%An interesting conclusion is that
According to this data, users appear to behave in two different modes: \emph{attention} and \emph{re-orientation}. Eye fixations happen in the attention mode, when users have ``locked in'' on a salient part of the scene, while movements to new salient regions happen in the re-orientation mode. Being able to identify such modes in real time, from either head or gaze movement, can be very useful for interactive applications. \update{Further results for the different conditions, and for the latitudinal direction, can be found in the supplement.} \update{Finally,} this data and findings can be leveraged for \textit{time-dependent} and \textit{head-based} saliency prediction, as we will show in Sections~\ref{subsec:tempSaliency} and~\ref{subsec:headPredictor}.

\begin{figure*}[!t]
	\centering
	\includegraphics[width=0.95\textwidth]{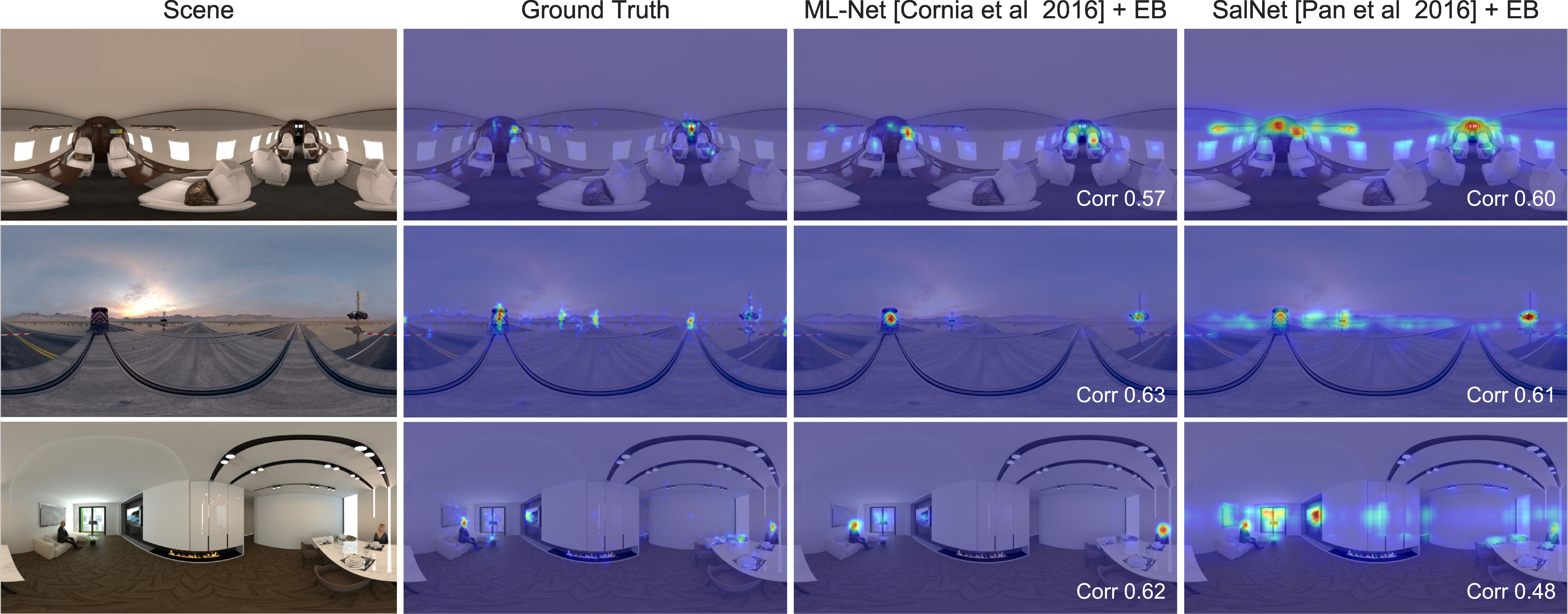}
	\caption{\label{fig:predicted_salmap} Saliency prediction for omni-directional stereo panoramas. Existing saliency predictors can be applied to spherical panoramas after they are projected onto a plane, here performed with the patch-based method described in the text. These methods tend to over-predict saliency near the poles. By multiplying the predicted saliency map by the longitudinal equator bias (EB) derived in the previous section, we achieve a good match between ground truth (center left) and predicted saliency (right). Note that this procedure could be applied to any saliency predictor; we chose two top-scoring predictors as an example.}
\end{figure*}

%Saliency prediction for omni-directional stereo panoramas can leverage existing saliency models. For this purpose, the target panorama (left) is divided into small patches (we use $60 \times 65^\circ$ per patch). Each of these patches can be represented with minimal perspective distortions and is processed with a saliency model of choice. The resulting patches are stitched together into a panorama and weighted by the latitudinal Laplacian function representing the \equ~bias (center right and right columns). Both quantitatively and qualitatively, in many cases this simple procedure achieves reasonable results compared to the ground truth saliency maps recorded with our gaze tracker in VR (second column).

%%%%%%%%%%%%%%%%%%%%%%%%%%%%%%%%%%%%%%%%%%%%%%%%%%%%%%%%%%%%%%%%%%%%%%%%%%%%%%%%%%%%%%%%%%%%%%%%%%%%%%%%%%%%%%%%%%%%%%%%%%%%%%%%%%%%%%%%%%%%%%%%%%%%%
\section{Predicting saliency in VR}
\label{sec:prediction}

In this section, we show how existing saliency prediction models can be adapted to VR using insights of our data analysis, such as the equator bias. Then, we ask whether the problem of time-dependent saliency prediction is a well-defined one that can be answered with sufficient confidence. Finally, we analyze how well head movement alone, for example captured with inertial sensors, can predict saliency without knowing the exact gaze direction. 

%%%%%%%%%%%%%%%%%%%%%%%%%%%%%%%%%%%%%%%%%%%%%%%%%%%%%%%%%%%%%%%%%%%%%%%%%%%%%%%%%%%%%%%%%%%%%%%%%%%%%%%%%%%%%%%%%%%%%%%%%%%%%%%%%%%%%%%%%%%%%%%%%%%%%
\subsection{Predicting saliency maps}
\label{subsec:predictors}

%%Saliency prediction is a well-explored topic and many existing models are evaluated by the MIT Saliency Benchmark~\cite{mitsaliencybenchmark}. Given an image, these models predict a saliency map in the form of a 2D probability distribution which characterizes the probability that a location is fixated within a two-second timeframe \belen{Check. Are we sure about this?}. The MIT benchmark and most existing prediction models, however, assume that users sit in front of a screen while observing the images -- ground truth data is collected by eye trackers recording precisely this behavior. 
%%%We argue that observing 360$^\circ$ panoramic image content in VR may be different enough to warrant closer investigation. 
%%% VR is different from conventional screens in that user snaturally use both head orientation and gaze to visually explore scenes. 
%%%We have shown before (Section~\ref{subsec:}) that low-level statistics in VR remain similar to those of traditional viewing. However, 
%%VR is different from traditional 2D viewing in that users naturally use both significant head movement and gaze to visually explore scenes, and we have seen that there is a strong relationship between them. In addition, kinematic constraints, such as sitting in a non-swivel chair, could affect user behavior. 
%%%\belen{It seems there may not be significant differences. We are waiting on exploration time to see... but we should cite here something of our study, since we have looked into this.}

Instead of learning VR saliency models from scratch, we ask whether existing models could be adopted to immersive applications. This would be ideal, because many saliency predictors for desktop viewing conditions already exist, and advances in that domain could be directly transferred to VR conditions. The fact that gaze statistics are closely related in VR and in traditional viewing (Section~\ref{subsec:lowlevel}) is indicative of the fact that existing saliency models may be adequate, at least to some extend, to VR. In this context, two primary challenges arise: (i) mapping a 360$^\circ$ panorama to a 2D image (the required input for existing models) distorts the content due to the projective mapping from sphere to plane; and (ii) head-gaze interaction may require special attention for saliency prediction in VR. We address both of these issues in the following.

%Before undertaking the task of building completely new saliency models for VR content, we should ask the question of whether existing models can be adapted to VR.
%%The datasets we recorded may contain an insufficient number of samples to train new data-driven behavioral models, but they may be sufficient \new{to evaluate existing saliency predictors} and to study the difference between viewing behavior in VR and with a conventional screen. 
%Ideally, we would like to leverage existing saliency prediction models for VR scenarios, and the data gathered in our study (Section~\ref{sec:Analysis}) offers insights into the feasibility of this, and into possible adaptations required. The fact that gaze statistics are similar in VR and in traditional viewing (Section~\ref{subsec:lowlevel}) sets a good basis for this hypothesis that existing saliency models may be applicable to VR (of course it is a necessary condition, but not a sufficient one). In this context, two primary challenges arise: (i) mapping a 360$^\circ$ panorama to a 2D image (the required input for existing models) distorts the content due to the projective mapping from sphere to plane; and (ii) head-gaze interaction may require special attention for saliency prediction in VR. We address both of these issues in the following. 

%%%%%%%%%%%%%%%%%%%%%%%%%%%%%%%%%%%%%%%%%%%%%%%%%%%%%%%%%%%%%%%%%
\subsubsection*{Which projection is best?}

\begin{figure}[t]
	\includegraphics[width=\columnwidth]{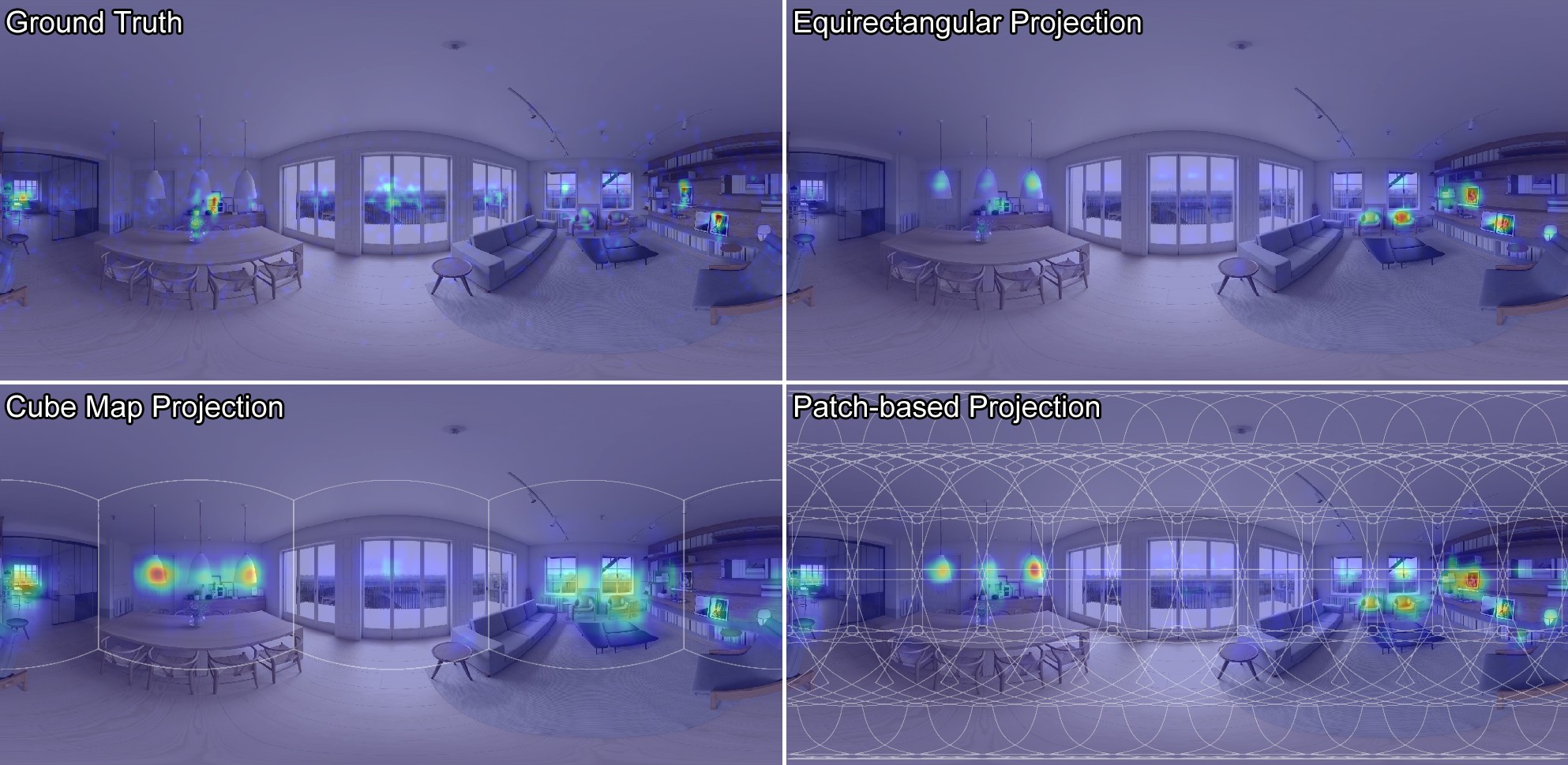}
	\caption{\label{fig:projection_comparison} Comparison of saliency prediction using different projections from sphere to plane. After applying the equator bias, all three projection methods result in comparable saliency maps for this example.}
\end{figure}

Before running a conventional saliency predictor on a spherical panorama or parts of it, the image has to be projected into a plane. Different projections would naturally result in different types of distortions that may affect the saliency predictor. For an equirectangular projection, for example, we expect large distortions near the poles. A cube map projection may result in discontinuities between some of the cube's faces. Alternatively, smaller patches can be extracted from the panorama, saliency prediction applied to each of them projected onto a plane, and the result stitched together and blended into a saliency panorama. The latter, patch-based approach would result in the least amount of geometrical distortions, but it is also the most computationally expensive approach and it gives up global context for the saliency prediction. 

In Figure~\ref{fig:projection_comparison} and Table~\ref{tab:projection_comparison_table} we compare saliency prediction using all three projection methods qualitatively and quantitatively. For each projection, we compute a saliency map using the state-of-the-art ML-Net saliency predictor~\cite{mlnet2016}, and then optionally multiply it by the latitudinal equator bias we derived in Section~\ref{subsec:bias}. \update{Figure~\ref{fig:projection_comparison} shows an example saliency map predicted on the three different sphere projections after applying the equator bias.} 
%\update{[Delete: Figure~\ref{fig:projection_comparison} shows that after applying our equator bias, possible artifacts of the equirectangular projection near the poles are not visible. Discontinuities are not observed in this particular example.]} 
We also compare the average CC score for all three projection methods and all 22~scenes in Table~\ref{tab:projection_comparison_table}. Quantitatively, saliency computed directly on the equirectangular projection with the equator bias applied not only performs best but it is also the fastest of the three approaches. \update{The benefit of applying the equator bias may be smaller for the equirectangular projection than for the other two projections, since the distortions at the poles may naturally lead to less saliency predicted at the poles than in the cube map and patch-based approaches.} %\update{[Delete: Likely, possible artifacts from distortions near the poles are mitigated by the equator bias.]}

{
\begin{table}
\scriptsize
	\centering
	\begin{tabular}{| l | l | l | l |}
		\cline{2-4}
		\multicolumn{1}{l|}{} & \multicolumn{1}{c|}{Equirectangular} & \multicolumn{1}{c|}{Cube Map} & \multicolumn{1}{c|}{Patch Based}  \\ \cline{1-4}
		$\!\!$ Without Equator Bias & $\mu  = $0.48  & $\mu  = $ 0.37 & $\mu  = $0.43	\\ \cline{1-4}
		With Equator Bias & \textbf{$\mu  = $0.50 } & $\mu  = $0.44  & $\mu  = $0.49 \\ \cline{1-4}
	\end{tabular}
	\caption{\label{tab:projection_comparison_table} Quantitative evaluation of three different projection methods with and without \equ~bias. We list the mean CC score for all 22~VR scenes used in this study. Applying the \equ~bias significantly improves the quality of all approaches. Distortions of the equirectangular projection near the poles do not affect saliency prediction as much as the shortcomings of other types of projection after the \equ~bias is applied.}
\end{table}
}

%
%First, we split the target panorama into several patches, each exhibiting minimal projective distortion so they can be processed separately by saliency predictors. Second, we weight the resulting saliency panorama by the \equ~bias derived in Section \ref{Analysis of the dataset} to account for longitudinal viewing preference. This simple approach is independent of any particular saliency predictor, and implemented as follows: 
%%
%\begin{algorithmic}[1]
%\State extract gnomonic projection patches from a panorama 
%\State apply a 2D saliency predictor to each patch 
%\State stitch patches back into a saliency panorama 
%\State weight saliency panorama by \equ~bias 
%\State normalize resulting saliency panorama  
%\end{algorithmic}

%%%%%%%%%%%%%%%%%%%%%%%%%%%%%%%%%%%%%%%%%%%%%%%%%%%%%%%%%%%%%%%%%
\subsubsection*{Which predictor is best?}

The fact that existing saliency predictors seem to apply to VR scenarios is important, because rapid progress is being made for saliency prediction with images and videos. Advances in those domains could directly improve saliency prediction in VR. Here, we further evaluate several different existing predictors both quantitatively and qualitatively. 

Table~\ref{tab:predictions} lists mean and standard deviation of the CC score for all 22~scenes in the VR condition, and for users exploring the same scenes in the desktop condition. These numbers allow us to analyze how good and how consistent across scenes a particular predictor is. We test the \equ~bias by itself as a baseline, as well as two of the highest-ranked models in the MIT benchmark where source code is available: ML-Net~\cite{mlnet2016} and SalNet~\cite{Pan2016}, together with the \equ~bias. We see that the two advanced models perform very similar and do much better than the \equ~bias alone. We also see that both of these models predict viewing behavior in the desktop condition better than for the VR condition. This makes sense, because the desktop condition is what these models were trained for originally. In Figure~\ref{fig:predicted_salmap} we also compare qualitatively the saliency maps of three scenes recorded under the VR condition (all scenes in the supplement). 

%Nevertheless, we believe that viewing behavior in VR may be driven by higher-level tasks and cognitive processes, possibly more so than observing conventional images. Current-generation saliency predictors do not take this into consideration. We hope that our dataset will be helpful for benchmarking future VR saliency predictors. Nevertheless, we believe that existing saliency predictors combined with the equator bias derived in this paper are a reasonable choice for saliency prediction in VR. The proposed approach is agnostic to the choice of saliency predictor for the ones we tested. 
%\belen{Not sure of leaving this. Discuss on absolute vs. relative saliency values.}
{
\begin{table}
\scriptsize
	\centering
	\begin{tabular}{| l | l | l | l |}
		\cline{2-4}
		\multicolumn{1}{l|}{} & \multicolumn{1}{c|}{EB} & \multicolumn{1}{c|}{ML-Net + EB} & \multicolumn{1}{c|}{SalNet + EB}  \\ \cline{1-4}
		$\!\!$ VR $\!\!$ & $\! \! \mu \! = \! 0.34 \pm 0.13 \!\! $ & $\!\! \mu \! = \! \textbf{0.49} \pm 0.11 \!\!$ & $\!\! \mu \! = \! 0.47 \pm 0.13 \!\!$ 	\\ \cline{1-4}
		Desktop & $\! \! \mu \! = \! 0.37 \pm 0.11 \!\!$ & $\! \! \mu \! = \! \textbf{0.57} \pm 0.11 \!\!$ & $\! \! \mu \! = \! 0.52 \pm 0.12\!\!$  \\ \cline{1-4}
	\end{tabular}
	\caption{\label{tab:predictions} Quantitative comparison of predicted saliency maps using a simple \equ~bias (EB), and two state-of-the-art models together with the EB. Numbers show average mean and standard deviation of CC scores, for each scene, between prediction and ground truth recorded from users exploring 22~scenes in the VR and desktop conditions. The proposed patch-based method was used to predict the saliency maps for both predictors.}
\end{table}
}
%
% TABLE BELOW IS THE SAME AS ABOVE, ONLY WITH OLD FORMAT
%%%{
%%%\begin{table}
%%%\scriptsize
%%%	\centering
%%%	\begin{tabular}{| l | l | l | l |}
%%%		\cline{2-4}
%%%		\multicolumn{1}{l|}{} & \multicolumn{1}{c|}{Equator Bias (EB)} & \multicolumn{1}{c|}{ML-Net~\shortcite{mlnet2016} + EB} & \multicolumn{1}{c|}{SalNet~\shortcite{Pan2016} + EB}  \\ \cline{1-4}
%%%		$\!\!$ VR $\!\!$ & $\! \! \mu \! = \! 0.33, \sigma \! = \! 0.13 \!\! $ & $\!\! \mu \! = \! 0.45, \sigma \! = \! 0.11 \!\!$ & $\!\! \mu \! = \! \textbf{0.45}, \sigma \! = \! 0.13 \!\!$ 	\\ \cline{1-4}
%%%		Desktop & $\! \! \mu \! = \! 0.43, \sigma \! = \! 0.12 \!\!$ & $\! \! \mu \! = \! \textbf{0.58}, \sigma \! = \! 0.11 \!\!$ & $\! \! \mu \! = \! 0.57, \sigma \! = \!  0.12\!\!$  \\ \cline{1-4}
%%%	\end{tabular}
%%%	\caption{\label{tab:predictions} Quantitative comparison of predicted saliency maps using a simple \equ~bias, and two state-of-the-art models. Numbers show average mean and standard deviation of CC scores, for each scene, between prediction and ground truth recorded from users exploring 22~scenes in the VR and desktop conditions. \belen{Nrs. need updating.}}
%%%\end{table}
%%%}
%
\begin{figure}[t]
	\includegraphics[width=0.9\columnwidth]{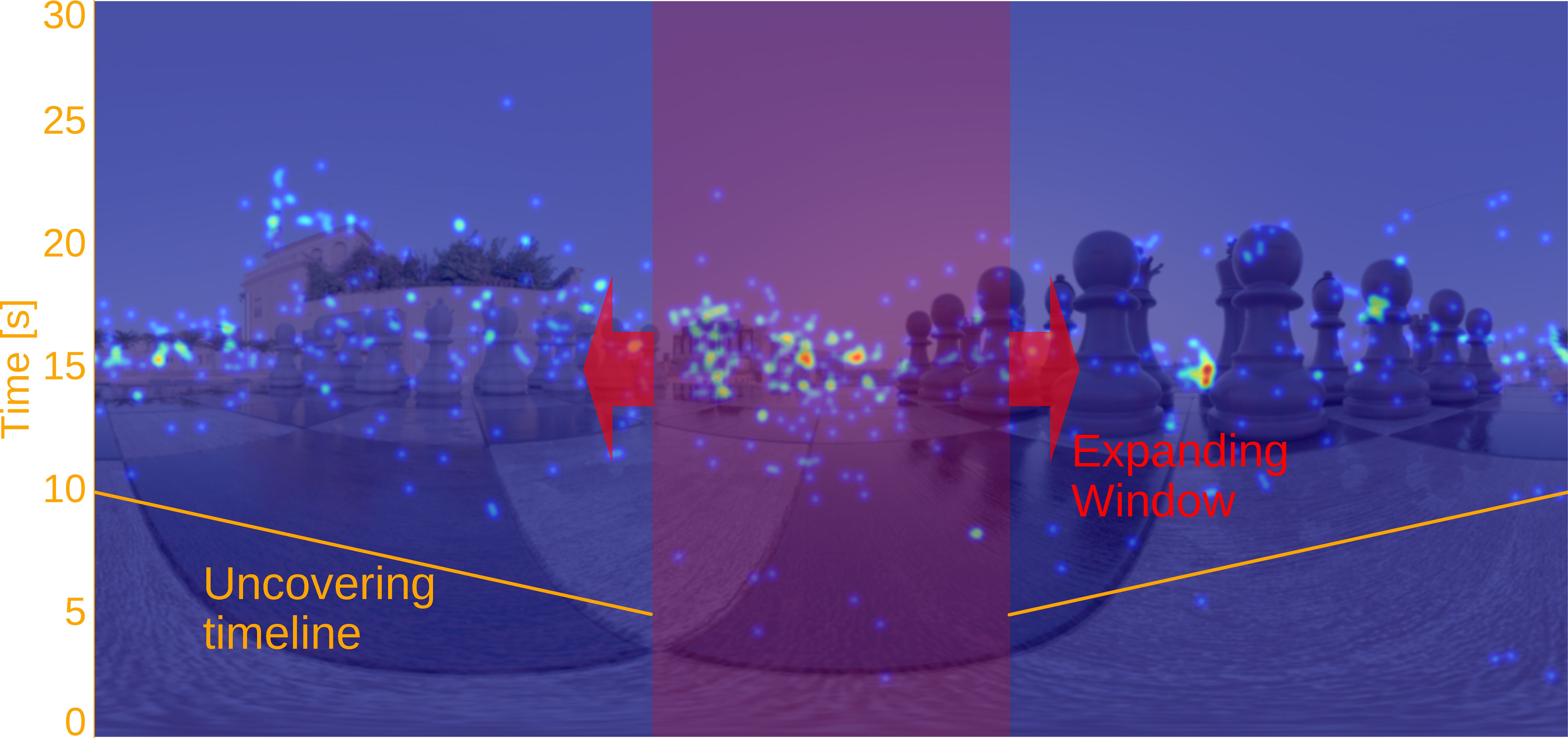}
	\caption{\label{fig:predicted_temp_salmap} Time-dependent saliency prediction by uncovering the converged saliency map with the average exploration speed determined in Section~\ref{sec:Analysis}.}
\end{figure}

%%%%%%%%%%%%%%%%%%%%%%%%%%%%%%%%%%%%%%%%%%%%%%%%%%%%%%%%%%%%%%%%%%%%%%%%%%%%%%%%%%%%%%%%%%%%%%%%%%%%%%%%%%%%%%%%%%%%%%%%%%%%%%%%%%%%%%%%%%%%%%%%%%%%%
\subsection{Can time-dependent saliency be predicted with sufficient confidence?}
\label{subsec:tempSaliency}

Virtual environments impose viewing conditions much different from those of conventional saliency prediction. Specifically, the question of temporal evolution arises: For users starting to explore the scene at a given starting point, is it possible to predict the probability that they fixate at specific coordinates at a time instant $t$? This problem is also closely related to scanpath prediction. We use data from Section~\ref{sec:Analysis} to build a simple baseline model for this problem: Figure~\ref{fig:ExpTime_Agreement} (right) shows an estimate for when users reach a certain longitude on average. We can thus model the time-dependent saliency map of a scene with an initially small window that grows larger over time to progressively uncover more of a converged (predicted or ground truth) saliency map. The part of the saliency map within this window is the currently active part, while the parts outside this window are set to zero. The left and right boundaries of the window are widened with the speed predicted in Figure~\ref{fig:ExpTime_Agreement} (right). 

Figure~\ref{fig:predicted_temp_salmap} visualizes this approach. We generate the time-dependent saliency maps for all 22 scenes and compare them with ground truth. We use the fully-converged saliency map as a baseline. The predicted, time-dependent saliency maps model the recorded data better than the converged saliency map within the first 6 seconds. Subsequently, they perform slightly worse until the converged map is fully uncovered after about 10 seconds, and the model is thus identical to the baseline. Our simple time-dependent model achieves an average CC score of 0.57 over all scenes, viewports, and the first 10 seconds (uncovering the ground truth saliency map), while using the converged saliency map as a predictor yields a CC of just 0.47. 

Although this is useful as a first-order approximation for time-dependent saliency, there is still work ahead to adequately model time-dependent saliency over prolonged periods. In fact, due to the high inter-user variance of recorded scanpaths\footnote{ While converged saliency maps show a high inter-user agreement (Section~\ref{subsec:agreement}), this is not necessarily the case for scanpaths, and thus for time-dependent saliency.}, the problem of predicting time-dependent saliency maps may not be a well-defined one. Perhaps a real-time approach that would use head orientation measured by an inertial measurement unit (IMU) to predict where a specific user will look next could be more useful than trying to predict time-dependent saliency without any knowledge of a specific user. 
%We leave this investigation for future work. 

%%%%%%%%%%%%%%%%%%%%%%%%%%%%%%%%%%%%%%%%%%%%%%%%%%%%%%%%%%%%%%%%%%%%%%%%%%%%%%%%%%%%%%%%%%%%%%%%%%%%%%%%%%%%%%%%%%%%%%%%%%%%%%%%%%%%%%%%%%%%%%%%%%%%%
\subsection{Can head orientation be used for saliency prediction?}
\label{subsec:headPredictor}

The analysis in Section~\ref{sec:Analysis} indicates a strong correlation between head movement and gaze behavior in VR. In particular, Figure~\ref{fig:head_fixation_vor} (middle) shows that fixations usually occur with low head velocities (except for the vestibulo-ocular reflex). This insight suggests that an approximation of a saliency map may be obtained from the longitudinal head velocity alone, e.g. measured by an IMU, without the need for gaze tracking. 
%In particular, as Figure~\ref{fig:head_fixation_vor} (middle) shows, low \emph{longitudinal head velocity} is an indicator of \emph{fixations}. This suggests that saliency maps obtained by thresholding longitudinal head velocity can be used as reasonable approximations of gaze saliency maps. 

We validate this hypothesis by counting the number of measurements at pixel locations where the head speed falls below a threshold of 19.6 $^\circ/s$ for all experiments in the VR condition. We then blur this information with a Gaussian kernel of size 11.7$^\circ$ of visual angle, to take into account the mean eye offset while fixating (Figure~\ref{fig:head_fixation_vor}, right). Qualitative results are shown in the supplemental material.  For a quantitative analysis, we compute the CC score between these \textit{head saliency maps} and the ground truth and compared it with the results obtained from the predictors examined in Table~\ref{tab:predictions}. Our CC score of 0.50 places our approximation on par with the performance of both saliency predictors tested; this is a positive and interesting result, given the fact that no gaze information is used at all. Head saliency maps could therefore become a valuable tool to analyze the approximate regions that users attend to from IMU data alone, without the need for additional eye-tracking hardware. 
%with 74 other state-of-the-art saliency predictors from the MIT Saliency Benchmark~\cite{mitsaliencybenchmark}. Our CC score of 0.52  places our approximation in the top 21\% (only ten of the 74 methods score a CC $>0.6$, while 28 score under 0.4); this is a positive and interesting result, given the fact that no gaze information is used at all. Head saliency maps could therefore become a valuable tool to analyze the approximate regions that users attend to from IMU data alone, without the need for additional eye-tracking hardware. 

%%%%%%%%%%%%%%%%%%%%%%%%%%%%%%%%%%%%%%%%%%%%%%%%%%%%%%%%%%%%%%%%%%%%%%%%%%%%%%%%%%%%%%%%%%%%%%%%%%%%%%%%%%%%%%%%%%%%%%%%%%%%%%%%%%%%%%%%%%%%%%%%%%%%%
\section{Applications}
\label{sec:applications}

In this section, we outline several applications for VR saliency prediction. Rather than evaluating each of the applications in detail and comparing extensively to potentially related techniques, the goal of this section is to highlight the importance and utility of saliency prediction in VR for a range of applications with the purpose of stimulating future work in this domain. 

%%%%%%%%%%%%%%%%%%%%%%%%%%%%%%%%%%%%%%%%%%%%%%%%%%%%%%%%%%%%%%%%%%%%%%%%%%%%%%%%%%%%%%%%%%%%%%%%%%%%%%%%%%%%%%%%%%%%%%%%%%%%%%%%%%%%%%%%%%%%%%%%%%%%%
\subsection{Automatic alignment of cuts in VR video}
\begin{figure}[t]
	\centering
	\includegraphics[width=0.9\columnwidth]{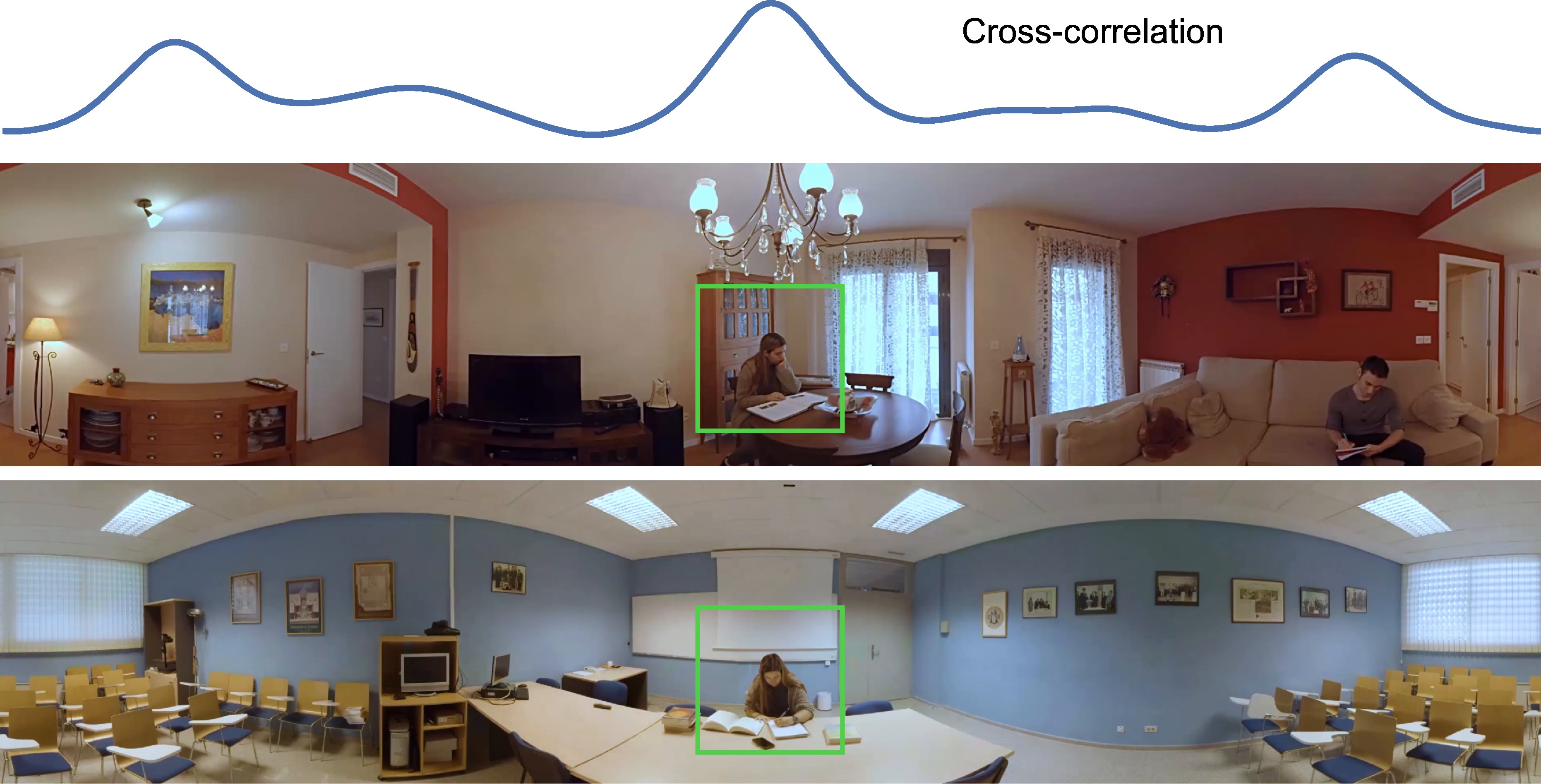}
	\caption{Automatic alignment of cuts in VR video. To align two video segments, we can maximize the correlation between the saliency maps of the last frame in the first segment and the first frame of the second one. The cross-correlation accounting for all horizontal shifts is shown on top of this example, which has been automatically aligned with the proposed algorithm.} 
	\label{fig:cut_alignment}
\end{figure}

How to place cuts in VR video is a question that was recently addressed by Serrano et al.~\cite{Serrano2017}. In a number of situations, alignment of the objects of interest before and after the cut is desirable.
The proposed saliency prediction facilitates automatic alignment of such cuts. We show in Figure~\ref{fig:cut_alignment} and in the supplemental video that predicted saliency maps can be used to align VR video before and after a cut by shifting the cuts in the longitudinal direction such that the Pearson CC of the predicted saliency maps is maximized. We use the 72 scenes provided by Serrano et al.~\cite{Serrano2017}, which were manually aligned to overlapping regions of interest (ROI) before and after a cut - however, for many of these scenes, several meaningful alignments are possible. Further, in some there are multiple ROIs, and thus multiple meaningful alignments possible. We predict saliency maps before and after the cut using the predictor described as performing best in Section~\ref{subsec:predictors} (i.e., ML-Net with \equ~bias on equirectangular projection), and then shift the saliency map after the cut with respect to the saliency map before the cut such as to maximize the Pearson correlation. 
For the scenes with one ROI visible before and after the cut, the median error of our method with respect to the manually aligned results is $2.11^\circ$, which mildly increases to $9.14^\circ$ if we include the scenes with two ROIs in the same field of view.
%Our method yields alignments that, in median, differ from the hand-aligned versions by $42.4^\circ$. 
Qualitative analysis shows that the alignments are meaningful and succeed to align salient regions, however, performance is strongly dependent on the quality of the saliency predictor used. This indicates that saliency-based automatic alignment of video cuts is a useful way to guide users when editing VR videos, suggesting good initial alternatives, but it may not be able to completely replace user interaction. Full alignment results can be found in the supplemental.

%For the 72~scenes provided by Serrano et al.~\shortcite{Serrano2017}, we obtain a median alignment error of $42^\circ$ between their manually-aligned cuts and our automatic alignment. This error is rather large and dominated by alignment of scenes that have multiple distinct ROIs; the algorithm aligns those with the highest saliency correlation whereas a user may manually select an alignment based on higher-level objectives, such as the story. Thus, saliency-based automatic alignment of video cuts is useful as a way to guide users in selecting cut alignments, but it may not be able to completely replace user interaction. 

%Visual inspection confirms that for most of the scenes, our algorithm correctly aligns the ROIs. Failure cases arise when the saliency predictor is inaccurate; an example of this can be seen in Figure~\ref{fig:cut_alignment} (bottom), in which, before the cut, the predictor erroneously detects a higher saliency in the picture than in the person coming through the door, possibly due to their dark clothing. Thus, saliency-based automatic alignment of video cuts works well in most scenarios with the simple method discussed here. Yet, an extensive evaluation of this application and improving robustness are a promising avenue of future research.

%%%%%%%%%%%%%%%%%%%%%%%%%%%%%%%%%%%%%%%%%%%%%%%%%%%%%%%%%%%%%%%%%%%%%%%%%%%%%%%%%%%%%%%%%%%%%%%%%%%%%%%%%%%%%%%%%%%%%%%%%%%%%%%%%%%%%%%%%%%%%%%%%%%%%
\subsection{Panorama thumbnails}

\begin{figure}[t]
    \centering
    \includegraphics[width=0.9\columnwidth]{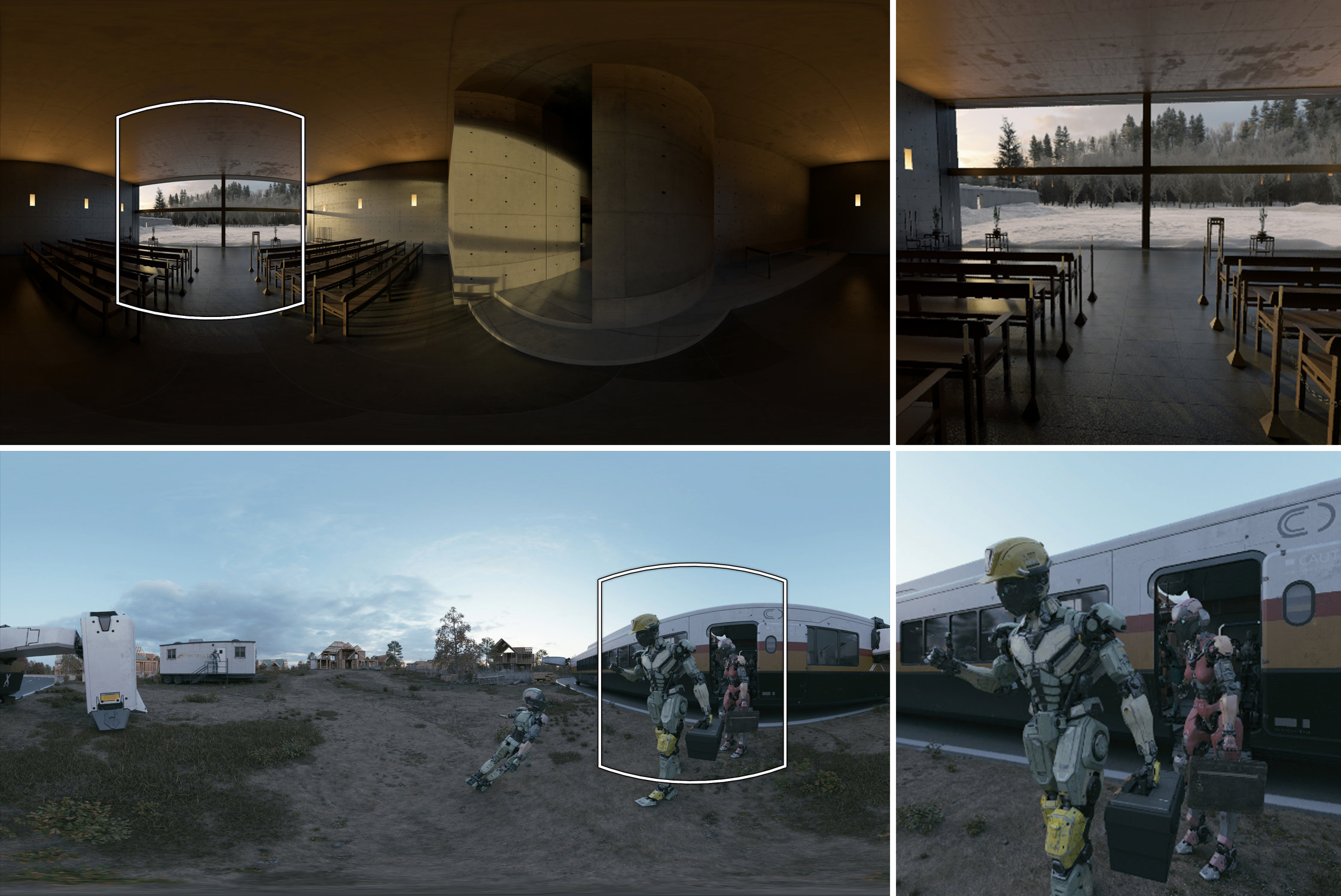}
    \caption{Automatic panorama thumbnail generation. The most salient regions of a panorama can be extracted to serve as a representative preview of the entire scene.} 
    \label{fig:thumbnails}
\end{figure}

Extracting a small viewport that is representative of a panorama may be helpful as a preview or thumbnail. However, VR panoramas cover the full sphere and most of the content may not be salient at all. To extract a thumbnail that remains representative of a scene in more commonly used image formats and at lower resolutions, we propose to extract the gnomonic patch of the panorama that maximizes saliency within. To this end, we compute the saliency map of the entire panorama as discussed in Section~\ref{subsec:predictors}. Then, we use an exhaustive search for the subregion with a fixed, user-defined field of view, that maximizes the integrated saliency within its gnomonic projection. A 2D Gaussian weighting function is applied to the saliency values within each patch before integration to favor patches that center the most salient objects. While this is an intuitive approach, it is also an effective one. Results are shown in Figure~\ref{fig:thumbnails} and, for all 22~scenes, in the supplemental material. Note that this approach to thumbnail generation is also closely related to techniques for gaze-based photo cropping~\cite{Santella:2006}.

%The predicted saliency maps can be leveraged to determine which viewport of a scene best represents its contents. To find this viewport, we proceed in five steps:
%\begin{algorithmic}[1]
%	\State predict the saliency map of the panorama following \ref{sec:prediction}
%	\State extract gnomonic projection patches from the saliency map
%	\State multiply each patch with a two-dimensional gaussian normal function to assign stronger weight to salient regions in the center
%	\State select the most salient patch and its coordinates
%	\State extract the same gnomonic patch from the panorama
%\end{algorithmic}
%This yields the most representative part of the scene as determined by visual saliency. 

%%%%%%%%%%%%%%%%%%%%%%%%%%%%%%%%%%%%%%%%%%%%%%%%%%%%%%%%%%%%%%%%%%%%%%%%%%%%%%%%%%%%%%%%%%%%%%%%%%%%%%%%%%%%%%%%%%%%%%%%%%%%%%%%%%%%%%%%%%%%%%%%%%%%%
\subsection{Panorama video synopsis}

\begin{figure}[t]
    \centering
    \includegraphics[width=0.9\columnwidth]{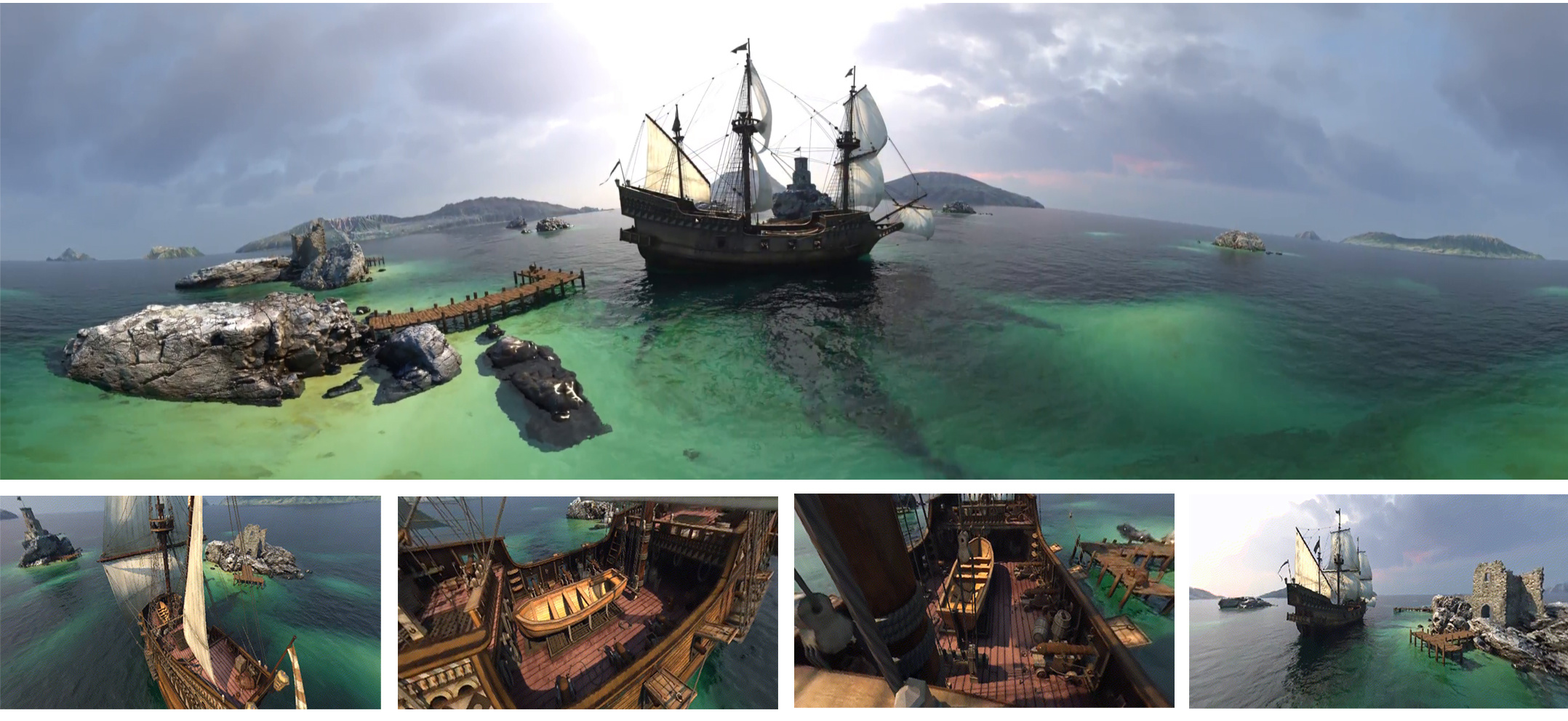}
    \caption{Automatic panorama video synopsis. Saliency prediction in VR videos can be used to create a short, stop-motion-like animation that summarizes the video. For this application, we predict saliency of each frame, extract a panorama thumbnail from one of the first video frames, and then search every $N^{th}$ frame for the window with highest saliency within a certain neighborhood of the last window.}
    \label{fig:video_summary}
\end{figure}

Automatically generating video synopses is an important and active area of research (e.g.,~\cite{RavAcha:2006}). Most recently, Su et al.~\cite{Su:pano2vid,Su:pano2vid2} introduced the problem of automatically extracting paths of a camera with a smaller field-of-view through 360$^\circ$ panorama videos, dubbed \emph{pano2vid}. Good saliency prediction for monoscopic and stereoscopic VR videos can help improve these and many other applications. Figure~\ref{fig:video_summary}, for example, shows an approach to combining video synopsis and \emph{pano2vid}. Here, we compute the saliency for each frame in a video and extracted the panorama thumbnail from the first frame as discussed in the last subsection. In subsequent frames, we search for the window in the panorama with the highest saliency that is close to the center of the last window. Neither the saliency prediction step nor this simple search procedure enforce strict temporal consistency, but the resulting panorama video synopsis works quite well (see supplemental video).

%%%%%%%%%%%%%%%%%%%%%%%%%%%%%%%%%%%%%%%%%%%%%%%%%%%%%%%%%%%%%%%%%%%%%%%%%%%%%%%%%%%%%%%%%%%%%%%%%%%%%%%%%%%%%%%%%%%%%%%%%%%%%%%%%%%%%%%%%%%%%%%%%%%%%
\subsection{Saliency-aware VR image compression}

Emerging VR image and video formats require substantially more bandwidth than conventional images and videos. Yet, low latency is even more critical in immersive environments than for desktop viewing scenarios. Thus, optimizing the bandwidth for VR video with advanced compression schemes is important and has become an active area of research~\cite{Yu:2015}. Inspired by saliency-aware video compression schemes~\cite{Hadizadeh:2014}, we test an intuitive approach to saliency-aware compression for omni-directional stereo panoramas. Specifically, we propose to maintain a higher resolution in more salient regions of the panorama. 

\begin{figure}[t]
    \centering
    \includegraphics[width=0.9\columnwidth]{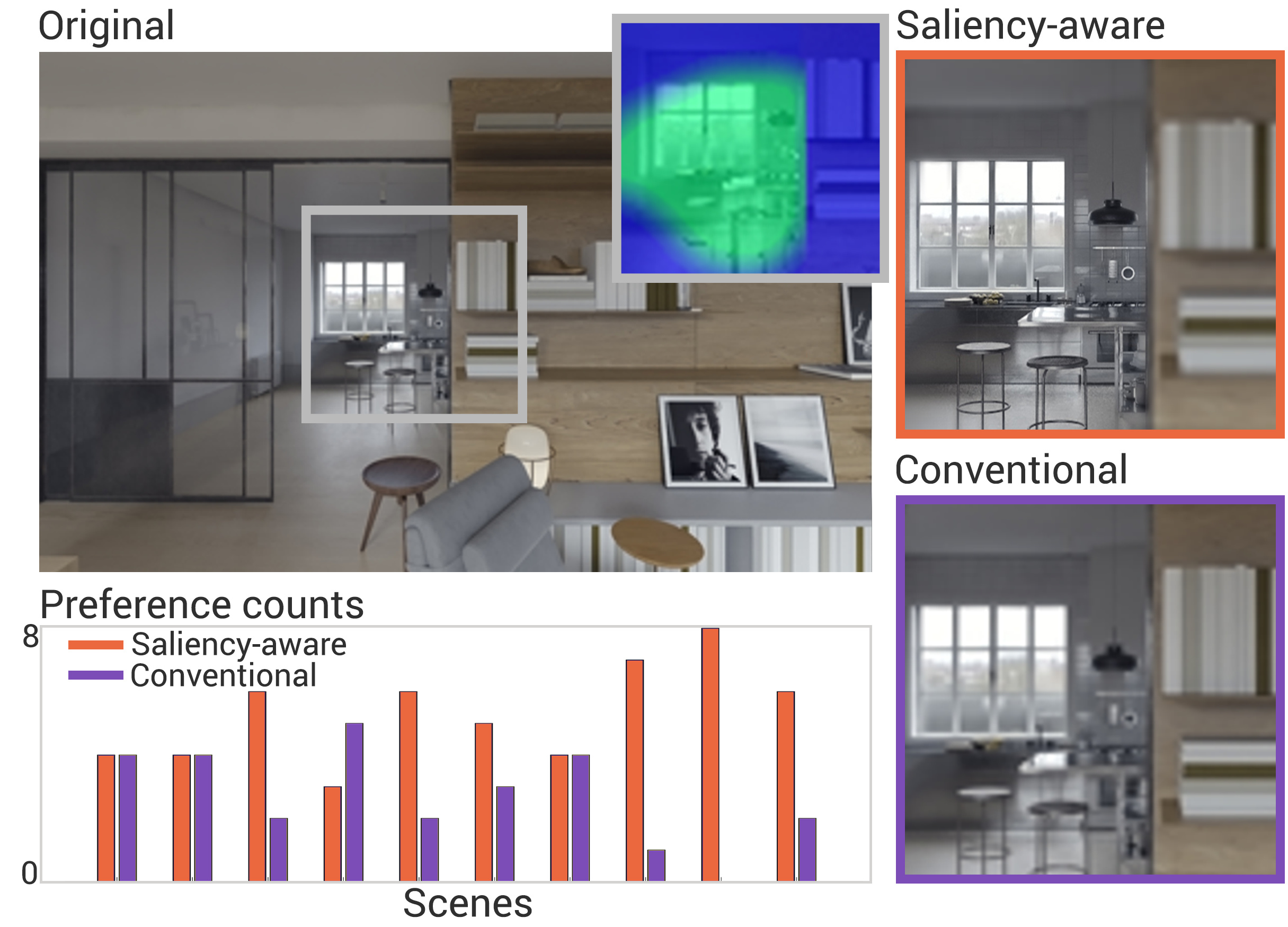}
	\caption{Saliency-aware panorama compression. \emph{Top left:} original, high-resolution region of the input panorama. Inset shows the compression map based on saliency information, where green indicates more salient regions. \emph{Right:} Close-ups showing the differences between saliency-aware compression and conventional downsampling. Note that salient regions retain a better quality in our compression, while non-salient regions get more degraded. \emph{Bottom left:} Preference counts for the ten scenes displayed during the user study.} 
  	\label{fig:compression}
\end{figure}

To evaluate potential benefits of saliency-aware panorama compression, we downsample a cube map representation of the omni-directional stereo panoramas with a bicubic filter by a factor of 6. We then upsample the low-resolution cube map and blend it with the 10\% most salient regions of the high-resolution panoramas. Overall, the compression ratio of the raw pixel count is thus 25\%. Figure~\ref{fig:compression} shows this saliency-aware compression for an example image.

%To this end, we downsample the whole panorama; doing this in equirectangular projections would cause progressively lower quality regions as they get closer to the top and bottom of the panorama, due to the equirectangular distortion. Instead, we perform bicubic downsampling separately on each of the twelve faces of the stereo panorama in cubemap projection. We then compute a \textit{compression map} by thresholding and smoothing saliency information. Last, for each face of the cubemap projection, the final compressed image is computed as a linear combination of the original and the low resolution images as dictated by the mask.  Figure~\ref{fig:compression} shows an example comparing our compression with respect to regular downsampling with an equal compression ratio of the $25\%$ of the original image.
%

To evaluate the proposed saliency-aware VR image compression, we carried out a pilot study to asses the perceived quality of saliency-aware compression when compared to regular downsampling for a comparable compression ratio. To this end, users were presented with ten randomized pairs of stereo panoramas, and they were asked to pick the one that had better quality in a two-alternative forced choice (2AFC) test. For each pair, we sequentially displayed the two panoramas in randomized order, with a blank frame of $0.75$ seconds between the two alternatives~\cite{Patney2016}. A total of eight users participated in the study, all reported normal or corrected-to-normal vision. The results of the study are shown in Figure~\ref{fig:compression} (bottom left). Saliency-aware compression was preferred for most scenes, and performed worse in only one scene. These preliminary results encourage future investigations of saliency-aware image and video compression for VR.

%Additionally we have carried out a perceptual study in order to asses the perceived quality of our compressed images when compared to regular downsampling. User were presented with \emph{ten} randomized pairs of stereo panoramas, and they were asked to pick the one that had better quality in a two-alternative forced choice (2AFC) test. For each pair, we sequentially displayed the two panoramas in randomized order, and with a blank frame of $0.75$ second between the two alternatives~\cite{Patney2016}. A total of eight users participated in the study, all reported normal or corrected to normal vision. The results of the study are shown in Figure~\ref{fig:compression} (bottom left). Our compressed panorama was preferred in most scenes, and similarly preferred in the rest of them.

\section{Discussion}
\label{Discussion}

In summary, we collect a dataset that includes gaze and head orientation for users observing omni-directional stereo panoramas in VR, both in a standing and in a seated condition. We also capture users observing the same scenes in a desktop scenario, exploring monoscopic panoramas with mouse-based interaction. The data encompasses 169 users in three different conditions, totaling 1980 head and gaze trajectories. All data will be publicly available.

The primary insights of our data analysis are: (1) gaze statistics and saliency in VR seem to be in good agreement with those of conventional displays; as a consequence, existing saliency predictors can be applied to VR using a few simple modifications described in this paper; (2) head and gaze interaction are coupled in VR viewing conditions -- we show that head orientation recorded by inertial sensors may be sufficient to predict saliency with reasonable accuracy without the need for costly eye trackers; (3) we can accurately predict time-dependent viewing behavior only within the first few seconds after being exposed to a new scene but not for longer periods of time due to the high inter-user variance; (4) the distribution of salient regions in the scene has a significant impact on how viewers explore a scene: the fewer salient regions, the faster user attention gets directed towards any of them and the more concentrated their attention is; (5) we observe two distinct viewing modes: attention and re-orientation, potentially distinguishable via head or gaze movement in real time and thus useful for interactive applications.

These insights could have a direct impact on a range of common tasks in VR. We outline a number of applications, such as panorama thumbnail generation, panorama video synopsis, automatically placing cuts in VR video, and saliency-aware compression. These applications show the potential that saliency has for emerging VR systems and we hope to inspire further research in this domain.

%: adaptive compression and streaming of omni-directional stereo content (see also~\cite{Yu:2015}), for example, can benefit from an understanding of what people are likely to look at in VR; while placing cuts in dynamic cinematic VR content adaptively in runtime can ensure that users see what the artists would like them to see. Additional applications of saliency prediction in VR that we show include panorama thumbnail generation, or panorama video synopsis. Further, placing information or advertisement into virtual environments could be more effective with a deeper understanding of saliency in VR. Finally, VR allows for unprecedented types of behavioral data to be collected and analyzed, which would be useful for basic cognitive science experiments and for learning behavior.

\paragraph{Future Work}
%The collected dataset of 1981 head and gaze trajectories for 22 omni-directional stereo panoramas could potentially allow predictive models for head and gaze trajectories to be learned. However, the number of samples may still be insufficient to train robust data-driven behavioral models, so future work could involve gathering more data for this purpose. It would be interesting to explore how such models could improve low-cost but imprecise gaze sensors, such as electrooculograms. In any case the data is enough to evaluate existing gaze or saliency predictors, an can thus serve as a benchmark for future testing. The data can be further exploited to study differences between viewing behavior in VR and on traditional screens. Finally, future work could extend the data collection and analysis to videos and multimodal experiences that include audio. 
Many potential avenues of future work exist. 
\update{We did not use a 3D display or mobile device since we wanted to closely resemble the most ÒstandardÓ viewing condition (regular monitor or laptop). Alternative viewing devices could be interesting for future work. Nevertheless, one of our goals is to analyze whether viewing behavior using regular desktop screens is similar to using a HMD, and our analysis seems to support this hypothesis. We believe this is an important insight, since it could enable future work to collect large saliency datasets for ODS maps without the need for HMDs equipped with eye trackers. }

Predicting gaze scanpaths of observers when freely exploring a VR panorama would be very interesting in many fields, including vision, cognition, and of course, any VR-related application. Since the seminal work of Koch and Ulman~\cite{Koch1987}, many researchers have proposed models of human gaze when viewing regular 2D images on conventional displays (e.g.,~\cite{LeMeur2015, Wang2011, Hacisalihzade1992, Boccignone2004}). An important element to derive such models is gaze statistics, and whether those found in our VR setup are comparable to the ones reported for traditional viewing conditions; this would inform to what extent we can use existing gaze predictors in VR applications, or be useful as priors in the development of new predictors. Our data can be of particular interest to build gaze predictors using just head movement as input, since head position is much cheaper to obtain than actual gaze data. 

Our data may still be insufficient to train robust data-driven behavioral models; we hope that making our scenes and code available will help gather more data for this purpose. \update{We also hope it will be a basis for people to further explore other scenarios, such as dynamic or interactive scenes, the influence of the task, or the presence of motion parallax, etc.; these future studies could leverage our methodology and metrics, and build upon them for the specific particularities of their scenarios.} It would be interesting to explore how behavioral models could improve low-cost but imprecise gaze sensors, such as electrooculograms. Future work could also incorporate temporal consistency for saliency prediction in videos, or extend it to multimodal experiences that include audio.

\section{Acknowledgements}
The authors would like to thank Jaime Ruiz-Borau for support with experiments. This research has been partially funded by an ERC Consolidator Grant (project CHAMELEON), the Spanish Ministry of Economy and Competitiveness (projects TIN2016-78753- P, TIN2016-79710-P, and TIN2014-61696-EXP), and the NSF/Intel Partnership on Visual and Experiential Computing (NSF IIS 1539120). Ana Serrano was supported by an FPI grant from the Spanish Ministry of Economy and Competitiveness. Gordon Wetzstein was supported by a Terman Faculty Fellowship and an Okawa Research Grant. We thank the following artists, photographers, and studios who generously contributed their omni-directional stereo panoramas for this study: Dabarti CGI Studio, Attu Studio, Estudio Eter, White Crow Studios, Steelblue, Blackhaus Studio, immortal-arts, Chaos Group, Felix Dodd, Kevin Margo, Aldo Garcia, Bertrand Benoit, Jason Buchheim, Prof. Robert Kooima, Tom Isaksen (Charakter Ink.), Victor Abramovskiy (RSTR.tv).

\bibliographystyle{abbrv-doi}

\bibliography{biblio}
\end{document}